\documentclass[10pt,twocolumn,letterpaper]{article}

\usepackage[pagenumbers]{cvpr}   %

\usepackage{cuted} %

\usepackage{currfile} %

\usepackage{caption} %
\usepackage{microtype} %

\usepackage{booktabs} %
\usepackage{multirow}
\newcommand{\method}{ScreenProof}
\newcommand{\judge}{ProofJudge}

\usepackage{colortbl}
\definecolor{tband}{HTML}{F3F0FD}
\definecolor{tacc}{HTML}{0E9F6E}
\newcommand{\mlogo}[1]{\raisebox{-0.14em}{\includegraphics[height=0.95em]{fig/logo_#1.png}}\hspace{0.3em}}
\newcommand{\mGF}{\mlogo{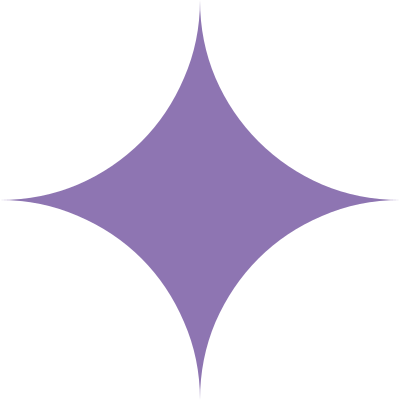}Gemini 3.8 Flash}
\newcommand{\mGP}{\mlogo{gemini}Gemini 3.1 Pro}
\newcommand{\mGA}{\mlogo{gemini}Gemini 4 Argon}
\newcommand{\mGPT}{\mlogo{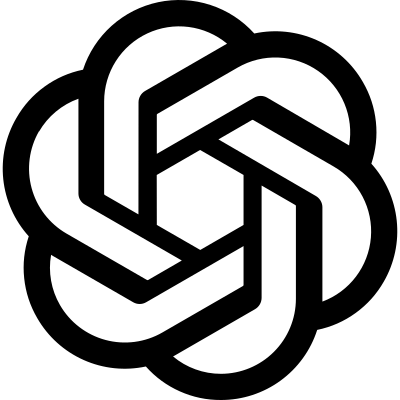}gpt-6-astra}
\newcommand{\mSOL}{\mlogo{openai}gpt-6.1-sol}

\newcommand{\mCL}{\mlogo{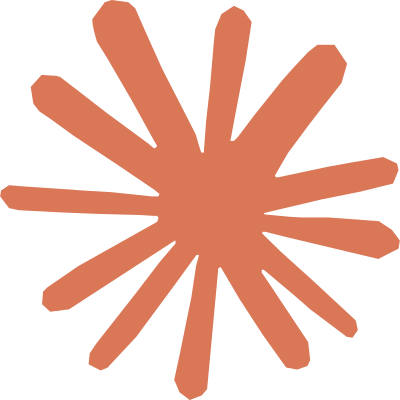}Claude Opus 5.5}
\newcommand{\mQW}{\mlogo{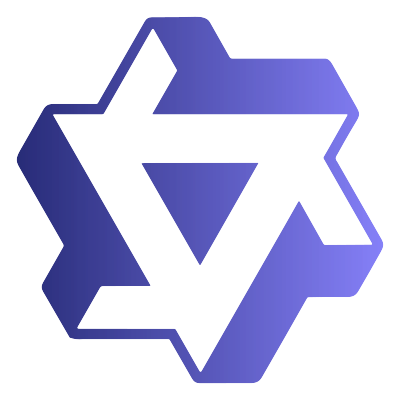}Qwen3.5-9B base}
\newcommand{\mRJ}[1]{\mlogo{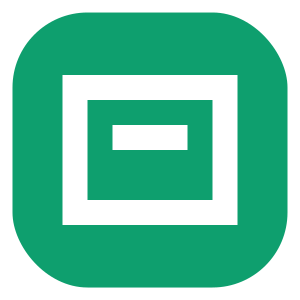}\judge\ #1}
\newcommand{\lo}[1]{\cellcolor{tacc!16}#1}
\newcommand{\tgroup}[2]{\rowcolor{tband}\multicolumn{#1}{@{}l}{\textbf{#2}}\\}
\newcommand{\tstyle}{\footnotesize\setlength{\tabcolsep}{4pt}\renewcommand{\arraystretch}{1.12}}

\newcommand{\RJFone}{0.48}
\newcommand{\RJPrec}{62\%}
\newcommand{\RJRealFA}{2\%}
\newcommand{\RJGenFA}{9\%}
\newcommand{\RJGenRec}{59\%}
\newcommand{\RJConRec}{25\%}

\newcommand{\RJIcon}{5\%}

\newcommand{\RJBestOpenFone}{0.15}

\newcommand{\RJCut}{0.01099}

\newcommand{\RJPages}{2,765}
\newcommand{\RJCleanPages}{607}
\newcommand{\RJCleanReal}{120}
\newcommand{\RJCleanGen}{487}
\newcommand{\RJGenTruth}{920}
\newcommand{\RJConTruth}{1,238}
\newcommand{\RJCtDisplay}{347}
\newcommand{\RJCtOverlap}{115}
\newcommand{\RJCtTwoselected}{111}
\newcommand{\RJCtType}{77}
\newcommand{\RJCtReorder}{65}
\newcommand{\RJCtContrast}{66}
\newcommand{\RJCtTarget}{48}
\newcommand{\RJCtDrop}{44}
\newcommand{\RJCtOverflow}{47}
\newcommand{\RJCtIcon}{188}
\newcommand{\RJCtColour}{179}
\newcommand{\RJCtMaterial}{123}
\newcommand{\RJCtRadius}{86}
\newcommand{\RJCtSwap}{662}
\newcommand{\RJCtLocal}{576}
\newcommand{\RJRawFone}{0.46}
\newcommand{\RJRawRealFA}{25\%}
\newcommand{\RJRawGenFA}{58\%}

\newcommand{\RJPrevFone}{0.40}
\newcommand{\RJPrevRealFA}{3\%}
\newcommand{\RJPrevGenFA}{8\%}

\newcommand{\RJVbFone}{0.33}
\newcommand{\RJVbRealFA}{32\%}
\newcommand{\RJVbGenFA}{46\%}
\newcommand{\RJHumanFone}{0.49}
\newcommand{\RJHumanArgonFone}{0.38}

\newcommand{\RJHumanDrop}{85}
\newcommand{\RJHumanChecked}{481}
\newcommand{\RJBootDiff}{+0.109}
\newcommand{\RJBootLo}{+0.086}
\newcommand{\RJBootHi}{+0.132}
\newcommand{\RJBootN}{10,000}
\newcommand{\RJOldFone}{0.31}
\newcommand{\RJOldRealFA}{4\%}
\newcommand{\RJOldGenFA}{11\%}
\newcommand{\RJOldArgonFone}{0.31}
\newcommand{\XFrIcon}{5--19\%}

\newcommand{\XApiGenRec}{31--49\%}
\newcommand{\XOpusFone}{0.31}

\newcommand{\XArgonFone}{0.37}
\newcommand{\XSolFone}{0.30}
\newcommand{\XArgonConRec}{41\%}

\newcommand{\XArgonPrec}{35\%}

\newcommand{\XArgonGenFA}{18\%}
\newcommand{\XArgonRealFA}{18\%}
\newcommand{\XArgonGenRec}{38\%}
\newcommand{\XSolGenRec}{45\%}
\newcommand{\XArgonRadius}{53\%}
\newcommand{\XSolRadius}{47\%}
\newcommand{\XArgonIcon}{15\%}
\newcommand{\XSolIcon}{11\%}
\newcommand{\XOpenFone}{0.00--0.15}
\newcommand{\XBaseFone}{0.04}
\newcommand{\XVsixFone}{0.21}

\newcommand{\XDtGenRec}{25--50\%}
\newcommand{\XDtBrandRec}{15--27\%}
\newcommand{\XDtPrec}{4--13\%}
\newcommand{\XDtFone}{0.07--0.17}
\newcommand{\XVfiveDtFone}{0.19}
\newcommand{\XArgonDtFone}{0.17}
\newcommand{\XVfiveDtBrand}{11\%}

\newcommand{\XOwFone}{0.72}
\newcommand{\XOwFA}{25\%}
\newcommand{\XOwAcc}{73\%}
\newcommand{\XIconVisN}{19}
\newcommand{\XIconBrandN}{43}

\newcommand{\XOpusIconBrandHit}{3}

\newcommand{\XApiRealFA}{18--62\%}
\newcommand{\XApiGenFA}{18--82\%}
\newcommand{\XApiTextAll}{47--72\%}
\newcommand{\XKnowMaterial}{91--99\%}
\newcommand{\XKnowMagic}{14--32\%}
\newcommand{\XFixNone}{95\%}
\newcommand{\XFixNoneCleanHarm}{12\%}
\newcommand{\XFixOracle}{95\%}
\newcommand{\XFixOracleCleanHarm}{0\%}
\newcommand{\XFixArgon}{55\%}

\newcommand{\XFixGateArgon}{72\%}
\newcommand{\XFixGateArgonSent}{56\%}

\newcommand{\XFixFaultPages}{85}
\newcommand{\XFixCleanPages}{40}
\newcommand{\XFixRJv}{25\%}

\newcommand{\XFixRJvCleanHarm}{0\%}
\newcommand{\XFixGateRJv}{28\%}
\newcommand{\XFixGateRJvSent}{22\%}
\newcommand{\XFixGateRJvClean}{5\%}
\newcommand{\XFixArgonCleanHarm}{15\%}
\newcommand{\XFixCostNone}{80}
\newcommand{\XFixCostGateArgon}{49}
\newcommand{\XFixCostGateRJv}{17}
\newcommand{\XFixCallCost}{0.080}

\newcommand{\XGateFone}{0.33}

\newcommand{\XGateRealFA}{1\%}
\newcommand{\XGateGenFA}{2\%}
\newcommand{\XGateCost}{1.88}

\newcommand{\XArgonRec}{40\%}
\newcommand{\XUnionOnly}{328}
\newcommand{\XUnionRec}{55\%}
\newcommand{\XUnionFone}{0.44}

\newcommand{\XUnionRealFA}{20\%}
\newcommand{\XUnionGenFA}{24\%}
\newcommand{\XSolGateSent}{42\%}
\newcommand{\XSolGateFone}{0.32}
\newcommand{\XSolGatePrec}{36\%}
\newcommand{\XSolGateRec}{28\%}
\newcommand{\XSolGateRealFA}{1\%}
\newcommand{\XSolGateGenFA}{4\%}
\newcommand{\XSolAloneFone}{0.30}
\newcommand{\XSolAlonePrec}{25\%}
\newcommand{\XSolAloneRec}{37\%}
\newcommand{\XSolAloneRealFA}{38\%}
\newcommand{\XSolAloneGenFA}{36\%}
\newcommand{\XGateCalls}{416}
\newcommand{\XTrRows}{17,285}
\newcommand{\XTrImages}{11,424}
\newcommand{\XTrPages}{990}
\newcommand{\XTrGenPages}{542}
\newcommand{\XTrRealPages}{448}
\newcommand{\XTrClean}{42\%}
\newcommand{\XTrPairs}{591}
\newcommand{\XTrRealRows}{790}
\newcommand{\XTrSpecs}{38}
\newcommand{\XTrOwRho}{0.58}
\newcommand{\XTrOwN}{21}
\newcommand{\XTrFaRho}{0.91}

\newcommand{\XRJSec}{1.0}
\newcommand{\XArgonSec}{6.1}
\newcommand{\XRJCost}{0.04}
\newcommand{\XArgonCost}{4.42}
\newcommand{\XOpenCost}{0.07--1.96}

\newcommand{\SVfivePrec}{20\%}
\newcommand{\SVfiveGenRec}{54\%}

\newcommand{\SVfiveMulti}{4\%}
\newcommand{\SSolMulti}{25\%}
\newcommand{\SSolMultiAll}{9}
\newcommand{\SSftFone}{0.18}
\newcommand{\SGrpoFone}{0.18--0.19}
\newcommand{\SGrpoBrand}{8--10\%}
\newcommand{\SMethodFone}{0.18--0.21}
\newcommand{\SRoneGenFrom}{57\%}
\newcommand{\SRoneGenTo}{40\%}
\newcommand{\SRoneFoneTo}{0.15}
\newcommand{\SFinalGenRec}{54\%}
\newcommand{\SFinalDtFone}{0.19}
\newcommand{\SSolSwapDt}{36\%}
\newcommand{\SAstraSwapDt}{33\%}
\newcommand{\SFinalSwapDt}{23\%}
\newcommand{\SSolStyleCI}{8--12\%}

\newcommand{\SVsixdFone}{0.33}
\newcommand{\SVsixdCon}{30\%}
\newcommand{\SVsixdGen}{55\%}
\newcommand{\SVsixdReal}{3\%}
\newcommand{\SValFone}{0.43}

\newcommand{\SValBestStep}{250}
\newcommand{\SValBestFone}{0.14}
\newcommand{\SValBestGenFA}{11\%}
\newcommand{\SValBestTest}{0.08}
\newcommand{\SValPrevTest}{0.25}

\newcommand{\SVsevencFone}{0.14}
\newcommand{\SVsevencCon}{2\%}
\newcommand{\SVsevencOwRec}{0\%}

\newcommand{\SVsevencValReal}{0\%}
\newcommand{\SVsevencValGen}{11\%}
\newcommand{\SEffortNoise}{0.00--0.08}
\newcommand{\BFaulty}{13}
\newcommand{\BPages}{75}
\newcommand{\BUncertain}{6}
\newcommand{\BAgree}{51}
\newcommand{\BControls}{0}
\newcommand{\BControlsN}{20}
\newcommand{\BKappa}{0.51--0.73}
\newcommand{\BRJGen}{7\%}
\newcommand{\BRJReal}{4\%}
\newcommand{\BArgonGen}{16\%}
\newcommand{\BSolGen}{39\%}
\newcommand{\BRJFone}{0.31}
\newcommand{\BArgonFone}{0.31}
\newcommand{\TFaults}{367}

\newcommand{\TArgonAll}{47\%}
\newcommand{\TArgonAllThirty}{51\%}
\newcommand{\TArgonCleanFA}{32\%}
\newcommand{\TArgonNegFA}{56\%}
\newcommand{\TArgonWord}{86\%}
\newcommand{\TArgonEllipsis}{81\%}
\newcommand{\TArgonLone}{22\%}
\newcommand{\TArgonTouch}{17\%}
\newcommand{\TArgonTouchWeak}{3\%}
\newcommand{\TArgonSpaceWeak}{0\%}
\newcommand{\TSolAll}{54\%}
\newcommand{\TSolAllThirty}{56\%}
\newcommand{\TSolCleanFA}{49\%}
\newcommand{\TSolNegFA}{68\%}
\newcommand{\TSolWord}{100\%}
\newcommand{\TSolEllipsis}{81\%}
\newcommand{\TSolLone}{25\%}
\newcommand{\TSolTouch}{27\%}
\newcommand{\TSolTouchWeak}{9\%}
\newcommand{\TSolSpaceWeak}{8\%}
\newcommand{\TOpusAll}{72\%}

\newcommand{\TOpusCleanFA}{80\%}
\newcommand{\TOpusNegFA}{96\%}

\newcommand{\TOpusTouch}{73\%}

\newcommand{\TRJAll}{43\%}

\newcommand{\TRJCleanFA}{15\%}
\newcommand{\TRJNegFA}{13\%}
\newcommand{\TRJWord}{94\%}
\newcommand{\TRJEllipsis}{96\%}

\newcommand{\TRJTouch}{5\%}

\newcommand{\CapVeGrpo}{0.37}
\newcommand{\CapVfGrpo}{0.41}
\newcommand{\CapVhPair}{0.42}
\newcommand{\CapSixtyFour}{0.45}
\newcommand{\CapVoteVal}{0.654}
\newcommand{\CapVoteTest}{0.535}
\newcommand{\CapVoteDiff}{+0.054}
\newcommand{\CapVoteLo}{+0.041}
\newcommand{\CapVoteHi}{+0.067}
\newcommand{\CapVoteRealFA}{8\%}
\newcommand{\CapVoteGenFA}{12\%}
\newcommand{\CapVoteOldGen}{18\%}
\newcommand{\CapVoteText}{21\%}
\newcommand{\CapVoteOw}{0.66}
\newcommand{\CapPairVal}{0.607}
\newcommand{\CapThreeEp}{0.619}
\newcommand{\CapVhVal}{0.612}
\newcommand{\CapRuns}{7}

\usepackage{placeins}

\definecolor{cvprblue}{rgb}{0.21,0.49,0.74}
\usepackage[pagebackref,breaklinks,colorlinks,allcolors=cvprblue]{hyperref}

\def\confName{arXiv}
\def\confYear{2026}

\title{ScreenProof: Where Is This Screen Wrong?\\ Finding UI Faults Without a Reference}

\author{%
Huiming Sun$^{1}$ \quad Yue Li$^{2}$ \quad Kangning Yang$^{1}$ \quad Ruineng Li$^{1}$ \quad Jie Cai$^{1}$\\
Lan Fu$^{1}$ \quad Yunwan Fu$^{3}$ \quad Jinlong Li$^{1}$ \quad Jinglin Shen$^{1}$\\[0.4em]
$^{1}$OPPO US Research Center \quad $^{2}$Cleveland State University \quad $^{3}$Shanghai Jiao Tong University\\
{\tt\small huiming.sun1@innopeaktech.com}
}

\begin{document}
\twocolumn[{%
\renewcommand\twocolumn[1][]{#1}%
\maketitle
\begin{center}
\centering
\includegraphics[width=\textwidth]{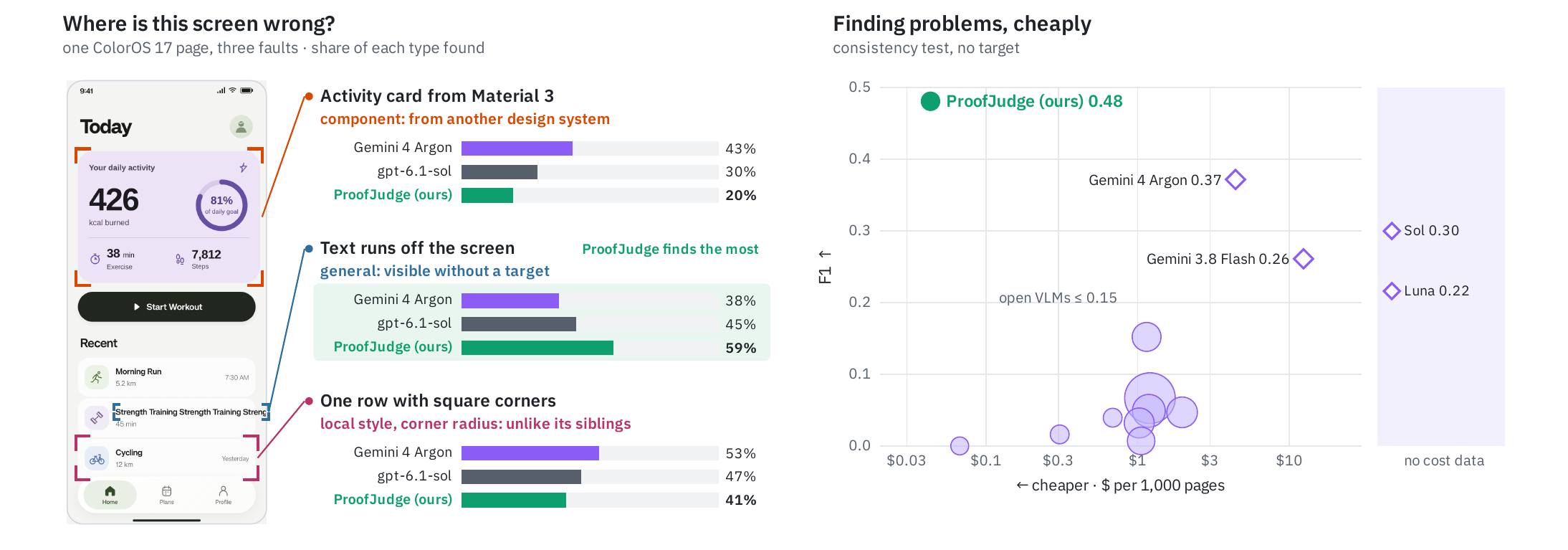}
\captionof{figure}{Where is this screen wrong, and which judge finds it (consistency test, \RJPages\ screenshots, no target). \textbf{Left:} one generated ColorOS 17 page with three faults the data engine made in its code: an activity card from a Material 3 Expressive page (a component from another design system), text that runs off the screen (general) and one list row with square corners (local style). Bars give the share of all test faults of that type each judge finds. \textbf{Right:} F1 against cost per 1,000 pages; point area is an open model's size, diamonds are API judges. \judge\ (9B) reaches \RJFone\ for \$\XRJCost, the highest F1 of all judges: above Gemini 4 Argon (\XArgonFone, \$\XArgonCost), gpt-6.1-sol (\XSolFone) and every open VLM (F1 at most \RJBestOpenFone, \$\XOpenCost). GPU cost assumes \$2 per GPU-hour; gpt-6.1-sol and GPT-6 Luna report no token counts.}
\label{fig:teaser}
\end{center}
\vspace{0.6em}
}]
\begin{abstract}
A phone screen can be wrong in ways that need no reference: a label that runs off the screen, letters that touch, a line that breaks one character before its end, one button with sharp corners among round ones.
We ask whether a model can box every such fault on a screenshot, given no target design and no reference image.
Such a critic could serve as a cheap first pass over generated pages before a designer reviews them, but only if it stays silent on clean pages.
\method\ poses this target-free critique over three families of faults: general faults, local style faults (one element breaks the style its page sets) and text layout faults (letter spacing, line height, line breaks).
A data engine makes each fault in page code and records the changed element's box, and hard negatives change every sibling alike.
No frontier judge is reliable: frontier models flag \XApiRealFA\ of clean real screenshots and \XApiGenFA\ of clean generated pages, find \XApiTextAll\ of text layout faults, and rarely find a foreign icon (\XFrIcon).
Our small critic \judge\ (Qwen3.5-9B with LoRA), trained on code-made pages with one to four faults each and keeping a finding only when it is confident, reaches F1 \RJFone, the highest of all judges (Gemini 4 Argon \XArgonFone; open models at most \RJBestOpenFone; \RJRawFone\ even without the confidence cut), and is the only fault-finding judge within 15\% false alarms on clean pages (\RJRealFA\ real, \RJGenFA\ generated), the property such a first pass needs, at about a hundredth of Argon's cost.
It finds the most general faults (\RJGenRec) but fewer local style faults than Argon (\RJConRec\ against \XArgonConRec) and fewer text layout faults than the API judges (\TRJAll; Gemini 4 Argon \TArgonAll).
Against a named brand the task is harder still: models know a design language about as well as its public documentation states it.
\end{abstract}

\section{Introduction}
\label{sec:intro}

\begin{figure*}[t]
\centering
\includegraphics[width=\textwidth]{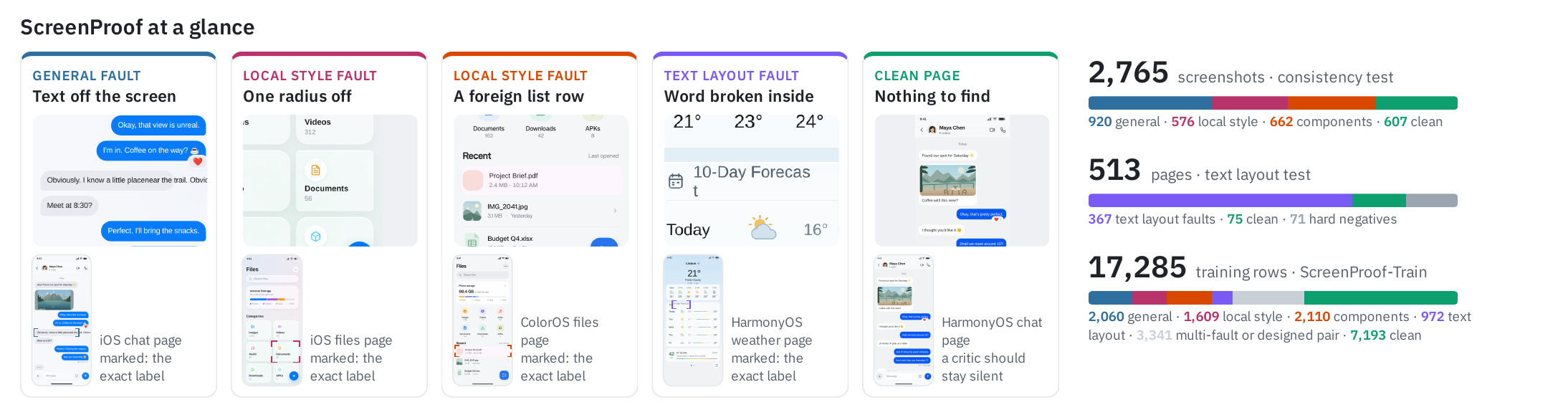}
\caption{\method\ at a glance. Left: one test example per fault family, each from another page: the faulty area enlarged on top, and the whole screen below with the data engine's exact label as corner marks; a clean page has nothing to find, and a critic should stay silent on it. Right: the two test sets and the released training set, split by fault family in the same colours (\cref{sec:task,sec:engine}). Every fault is made in page code, so its box is exact.}
\label{fig:overview}
\end{figure*}

Many faults on a screen need no reference to be seen.
Some are wrong in any design: a label that runs off the screen, two cards that overlap, a button too small to tap.
Some break the style the page itself sets: one button with sharp corners among round ones, one icon from another set, one component lifted from another design system.
And some are faults of text layout: letters that touch, lines that crowd, a word broken in the middle, a line that ends on a single character.
Designers see these at a glance, but a useful review must say \emph{where}: which element is wrong and why.
The question is newly pressing in generation.
Multimodal models now turn screenshots and text into front-end code~\cite{si2025design2code,yang2025ui2coden,xiao2026comuicoder,wu2026mobileforge}, and their pages place text worse than they copy it~\cite{si2025design2code,guo2026render}, while edits change parts nobody asked to change~\cite{xiao2025designbench,li2026mtweb2code}.
A generated page usually comes with no reference image and no specification, so the check must use the screenshot alone.
\emph{Such a critic has a direct use.} It could check every generated page before a designer sees it, pass the clean ones, and send back only the flagged ones with the element and the reason; the box can also guide a code fixer or a generator.
To save review time, the critic must stay silent on clean pages: one that flags most clean pages sends every page back to a human.

We seek one critic that boxes every such fault on a screenshot, given no target design and no reference. \method\ measures this and shows that no model can do it yet, and why.

\emph{Critics cannot tell a clean page from a faulty one.}
Frontier judges flag \XApiRealFA\ of clean real screenshots and \XApiGenFA\ of clean generated pages (\cref{tab:consistency}); none stays within 15\% on both.

\emph{One element off its own page often escapes every judge.}
Asked whether any element breaks the style set by the rest of the page, frontier judges find a changed colour or surface material most often, but catch a single foreign icon in only \XFrIcon\ of cases; many icon faults look normal on their page and are foreign only by a design language's conventions (\cref{fig:fault-recall}).

\emph{Text layout is found unevenly.}
Frontier judges find \XApiTextAll\ of text layout faults (\cref{tab:text-layout}): nearly every word broken in the middle, but few labels whose characters touch, and many pages where every sibling shares the same wide spacing are flagged.

\emph{Our critic addresses the first failure, false alarms.} Training narrows the gap without closing it: our small critic \judge\ (Qwen3.5-9B with LoRA, \cref{fig:rolejudge}), trained on pages with one to four faults each and keeping only confident findings, reaches the highest F1 of all judges (\RJFone; \RJRawFone\ without its confidence cut; Gemini 4 Argon \XArgonFone) at a fraction of the cost of an API judge (\cref{fig:teaser}), and is the only fault-finding judge that stays silent on at least 85\% of clean real and generated pages, the property a pre-filter needs. It recalls more general faults than any other judge (\RJGenRec; API judges \XApiGenRec), yet finds fewer local style faults than Gemini 4 Argon (\RJConRec\ vs.\ \XArgonConRec), few foreign icons (\RJIcon) and \TRJAll\ of text layout faults.

Exact labels are costly to annotate, so we synthesize them: a data engine injects faults in page code, recording exactly which element changed, and hard negatives change every sibling alike so that only a difference from the page counts.
An external benchmark of real screenshots (OwlEye) tests whether a critic's skill transfers beyond our pages.
A named brand makes the task harder still: models know a design language about as well as its public documentation states it (\cref{sec:findings}; \cref{sec:app-brand}).

\paragraph{Contributions.}
\begin{itemize}
\item \textbf{Benchmark.} \method: target-free critique of phone screenshots over general, local style and text layout faults, with exact boxes, clean pages and hard negatives. It is built from public sources and released in full. Brand tracks (knowledge, critique, design): \cref{sec:app-brand}.
\item \textbf{Data engine.} Faults are injected in page code, not pasted onto pixels, so the edit is the label: general faults, one element against its own page (radius, material, colour, icon, or a component from another design system), and one text element's spacing, line height or line breaks.
\item \textbf{Diagnosis and a baseline.} Why the task remains unsolved, measured on our tests and an external benchmark, and \judge, a lightweight critic trained on multi-fault pages, the highest F1 on our consistency test and the only judge within the false-alarm gate, at about \$\XRJCost\ per 1,000 pages; it shows what training a small critic does and does not fix.
\end{itemize}

\section{Related Work}
\label{sec:related}

\paragraph{VLMs as judges.}
FOCUS~\cite{focus2026} perturbs image-to-text and text-to-image outputs to measure how often VLM judges miss errors, using perturbations only to benchmark judges without covering UI.
Visual manipulation inflates LVLM judge scores~\cite{foolinglvlm2025}, rubrics aid aesthetic judges~\cite{jadhav2026rubric}, and MobileForge~\cite{wu2026mobileforge} admits a judge only if it ranks a hidden original first.
UI2Code$^\text{N}$~\cite{yang2025ui2coden} fine-tunes a judge for whole-page comparison.
Instead, we train a judge on exact-label UI faults to locate the faulty element.

\paragraph{Target-free critique and false alarms.}
A critic without a reference must decide both whether a page has a fault and where it is. Judge studies measure the first half: evaluators miss perturbed errors~\cite{focus2026} and can be pushed to inflated scores~\cite{foolinglvlm2025}, and UI defect benchmarks report recall on faulty screens~\cite{owleye,uilens2026}. How often a critic boxes something on a page with nothing wrong is rarely reported; we score it on clean pages and hard negatives, and it is where frontier judges fail.

\paragraph{UI quality and design evaluation.}
UI-Lens~\cite{uilens2026} evaluates general MLLMs on UI display defects with expert labels: nine MLLMs reach an average F1 of 22.19 on text overflow and 11.44 on text that differs across screens. OwlEye~\cite{owleye} detects display issues such as overlapping and missing text in human-built apps, and design-to-code benchmarks find that models copy text content well but place it badly (GPT-4o: 98.2 on text similarity, 85.5 on position)~\cite{si2025design2code}.
UICrit~\cite{duan2024uicrit} collects designer critiques for automated evaluation.
Design-system compliance of LLM-generated UI has been measured via specification checklists and an LLM judge~\cite{dscompliance2026chi}.
We are not aware of work that boxes elements inconsistent with their own page, and scores false alarms on clean pages and on hard negatives, without a reference.

\paragraph{Benchmarks for UI generation and critique.}
Design2Code~\cite{si2025design2code} scores screenshot-to-HTML by visual similarity, WebGen-V~\cite{wang2025webgenv} section-level visual design, DesignBench~\cite{xiao2025designbench} generation, editing and repair across frameworks, MobileForge~\cite{wu2026mobileforge} multi-screen apps by build, navigation and fidelity, and EvoGenUI-Bench~\cite{peng2026evogenui} multi-turn generative UI.
Closest to us, TaoD2C-Bench~\cite{shi2026taod2c} goes beyond visual fidelity with 97,652 expert annotations of required versus permitted implementation constraints on 2,861 production designs; whereas its constraints belong to one product's code and are checked in generated code, ours are faults of the rendered screen, checked from pixels alone.
For critique, WiserUI-Bench~\cite{jeon2026wiserui} asks which of two A/B-tested designs won, UXBench~\cite{mao2026uxbench} poses usability questions on mobile screenshots, a second UXBench~\cite{wang2026uxbench} scores critique actionability, and PerceptUI~\cite{bougie2026perceptui} simulates users.
None boxes reference-free faults of a single screen with exact labels and hard negatives.

\paragraph{UI understanding and grounding.}
Whereas UI datasets~\cite{deka2017rico,bunian2021vins}, screen models~\cite{baechler2024screenai,you2024ferretui} and GUI agents~\cite{hong2024cogagent,cheng2024seeclick,gou2025uground} locate elements matching a description, we locate elements that \emph{break} a design language.

\begin{table*}[!tp]
\centering\small
\caption{Positioning. Entries follow the papers' own descriptions.}\label{tab:related}
\resizebox{\textwidth}{!}{%
\begin{tabular}{@{}llccccc@{}}
\toprule
 & Kind & UI & Target & Locates elements & Names convention & Judge trained / admitted on faults \\
\midrule
Design2Code~\cite{si2025design2code} & generation benchmark & yes & one screenshot & block matching & -- & -- \\
UI2Code$^\text{N}$~\cite{yang2025ui2coden} & generation & yes & one screenshot & -- & -- & fine-tuned whole-page judge \\
ComUICoder~\cite{xiao2026comuicoder} & generation & yes & pages & per element (detector) & -- & -- \\
MobileForge~\cite{wu2026mobileforge} & generation benchmark & yes & screens & -- & -- & positive anchor only \\
MultiRef~\cite{multiref2025} & generation & -- & references with roles & -- & -- & -- \\
DS-Compliance~\cite{dscompliance2026chi} & generation study & yes & design-system spec & -- & checklist & -- \\
FOCUS~\cite{focus2026} & judge benchmark & -- & task output & -- & -- & perturbations, evaluation only \\
UI-Lens~\cite{uilens2026} & judge benchmark & yes & display quality & boxes & defect categories & -- \\
\midrule
\method\ + \judge\ (ours) & judge + benchmark & yes & none (named brand: \cref{sec:app-brand}) & boxes & yes & code-level faults \\
\bottomrule
\end{tabular}}
\end{table*}

\paragraph{UI generation and self-correction.}
Screenshot-to-code methods~\cite{wan2025dcgen,wu2025layoutcoder} and rendered-feedback loops~\cite{yang2025ui2coden,xiao2026comuicoder} reproduce a reference, while MultiRef~\cite{multiref2025} conditions generation on role-annotated references.
Generated pages place text worse than they copy it~\cite{si2025design2code,guo2026render}, and edits alter unrequested parts~\cite{xiao2025designbench,li2026mtweb2code}.
Since self-correction without external feedback is unreliable~\cite{madaan2023selfrefine,huang2024cannotselfcorrect}, our fix test gives the fixer a judge's findings, a judge's gate or no hint (\cref{sec:fix}).
\Cref{tab:related} compares these works feature by feature.

\section{\method}
\label{sec:task}

\begin{table}[t]
\centering\tstyle
\caption{Fault taxonomy. The main task gives no target; a finding is general, local style or text layout (brand faults, which need a named target: \cref{sec:app-brand}). \emph{Engine fault}: how the data engine makes it (\cref{sec:engine}).}\label{tab:taxonomy}
\begin{tabular}{@{}>{\raggedright\arraybackslash}p{0.19\columnwidth}>{\raggedright\arraybackslash}p{0.45\columnwidth}>{\raggedright\arraybackslash}p{0.28\columnwidth}@{}}
\toprule
Aspect & What goes wrong & Engine fault \\
\midrule
\tgroup{3}{General: visible without a design target}
Layout & overflow, overlap, order, missing item, two selected & each, in code \\
Readability & contrast, tap size, typeface & each, in code \\
Display & broken image, null value, needless ellipsis & each, in code \\
\midrule
\tgroup{3}{Local style: one element unlike the rest of its page}
Style & radius, material, accent, icon & one element \\
Component & a control from another design system & one transplant \\
\midrule
\tgroup{3}{Text layout: one text element}
Spacing & characters touch, spacing unlike siblings & one text element \\
Lines & lines touch, line height unlike siblings & one text element \\
Breaks & short label wraps, lone last word, word broken, needless ellipsis, CJK punctuation at line start & one text element \\
\bottomrule
\end{tabular}
\end{table}

\paragraph{Task.}
The input is a screenshot $S$ and nothing else: no target design, no reference image, no specification.
The output is a verdict $y\in\{\text{clean},\text{faulty}\}$ and a set of findings $\{(B_i,e_i,k_i,c_i)\}$: a box $B_i$ on $S$, the element type $e_i$, the fault family $k_i\in\{$general, local style, text layout$\}$ (\cref{tab:taxonomy}), and a short reason $c_i$.
For local style faults the page is its own reference: an element is wrong because it differs from its siblings, so no brand knowledge is needed.
Because only pixels are given, the task is independent of how the screen was built: $S$ may come from a web page, a native app or a design-tool export; the task definition does not depend on which.

\paragraph{Test sets.}
The consistency test (\cref{fig:overview}) holds \RJPages\ screenshots, \RJCleanPages\ of them clean (\RJCleanReal\ real, \RJCleanGen\ generated), with \RJGenTruth\ general faults (display faults included) and \RJConTruth\ local style faults (transplanted components included), one per faulty page; the text layout test holds 367 faults on held-out page requirements, 75 clean pages and 71 hard negatives in which every sibling changes alike (\cref{sec:engine}).
Every page in both sets is scored without a target and with the same prompt.

\paragraph{Metrics.}
We report localization precision and recall, where findings and labelled elements are matched one-to-one at box IoU $\ge 0.5$ (so a small box inside a large faulty element does not count), alongside scores at IoU $\ge 0.3$ (\cref{tab:consistency}; \cref{tab:iou}), box F1, and the share of clean pages and hard negatives a judge flags.
Results are reported per family and type.

\paragraph{Data.}
\method\ comprises pages collected from the web: phone screenshots from the vendors' own pages (product pages, user guides, design guidelines), screens from public UI datasets, pages built with open-source UI libraries, generated base pages in eight phone design languages, and the fault copies the data engine makes from them.
It is released in full, images included, with the source, date and terms of each item.
\Cref{sec:app-data} describes the collection, with sample screenshots and counts (\cref{fig:dataset,fig:datasetcounts}).

\section{Data Engine}
\label{sec:engine}

\begin{figure*}[t]
\centering
\includegraphics[width=0.78\textwidth]{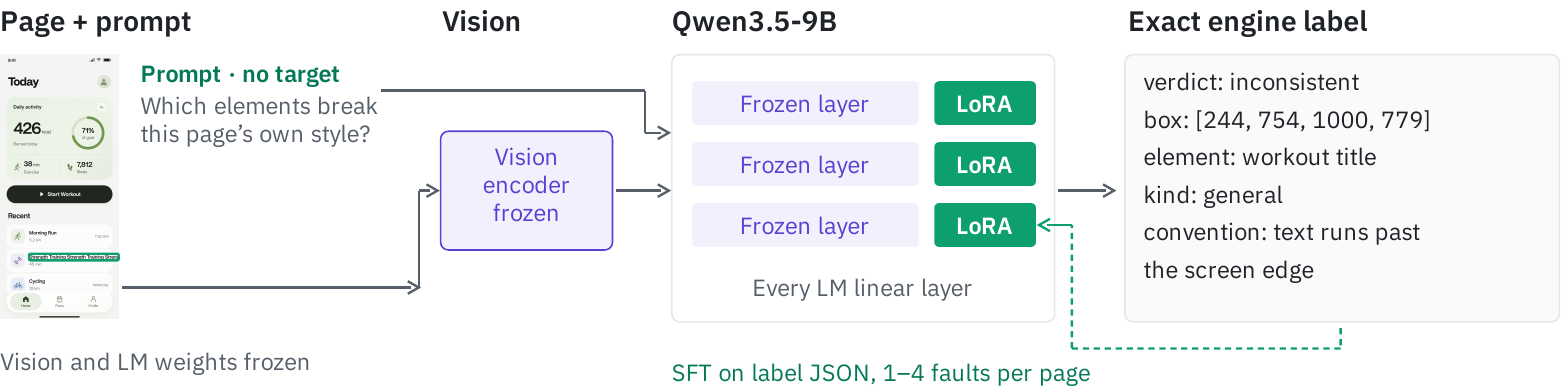}
\caption{Overview of \judge. A data-engine page and prompt pass through Qwen3.5-9B with its vision encoder and weights frozen; one LoRA on every linear layer of the language model is trained by SFT on the label JSON across pages with one to four faults each. The prompt gives no target and asks which elements break the page's own style. The label is exact: the engine made the title overflow, a general fault under any design.}
\label{fig:rolejudge}
\end{figure*}

\begin{table}[t]
\centering\tstyle
\caption{ScreenProof-Train: training rows and images by kind. A row is an image with its labels; an image can appear in several rows.}\label{tab:train}
\begin{tabular}{@{}lrr@{}}
\toprule
Kind & Rows & Images \\
\midrule
Clean page & 7,193 & 2,605 \\
Designed difference (faulted twin) & 630 & 591 \\
General fault & 1,542 & 1,427 \\
Local style fault & 1,609 & 1,505 \\
Transplanted component & 2,110 & 1,484 \\
Text layout fault & 972 & 847 \\
Page structure (dropped item, reordered sections) & 518 & 444 \\
Two to four faults of the kinds above & 2,711 & 2,521 \\
\midrule
All & 17,285 & 11,424 \\
\bottomrule
\end{tabular}
\end{table}

The engine operates on HTML pages. HTML lets us make exactly one change, read the element's rendered box from the DOM, and re-render the page identically on any machine; a judge sees only the screenshot, though judges still flag generated and real screens at different rates (\cref{tab:consistency}).
The engine turns a clean page into copies with known faults. Because each fault is injected in code, the engine knows the changed element, its rendered box and its kind in \cref{tab:taxonomy}; these form the label. Copies whose change is not visible on the first screen are discarded.

\paragraph{Base pages.}
For each of 14 page requirements, a generator writes an English base page in each of eight phone design languages and marks each component root with its kind (switch, tab bar, top bar, button, list row, search, chip, segmented control, slider, dialog, sheet, floating button), in light and dark mode (224 planned). Three requirements per design language are held out for testing.
A base page serves as a clean page only if two judges find it on-spec item by item; 80\% of the 223 generated pages pass (\cref{sec:app-brand}).

\paragraph{General faults.}
Display faults (a cropped or missing image, a null value, a needless ellipsis), overlap, text overflow, low text contrast and a shrunk tap target are visible in any design; a dropped item, reordered sections, two selected items in a single-choice group and a block set in another typeface are general faults that break the structure the page sets. The consistency test holds \RJGenTruth\ such faults: display \RJCtDisplay, overlap \RJCtOverlap, two selected \RJCtTwoselected, typeface \RJCtType, contrast \RJCtContrast, reordering \RJCtReorder, tap target \RJCtTarget, overflow \RJCtOverflow\ and dropped item \RJCtDrop.

\paragraph{Local style faults.}
One element of a kind that appears at least twice on the page takes another corner radius, material, accent or icon, while its siblings keep the page's own value; for example, one button's radius drops from 28 to 9\,px while all cards keep 28 (\cref{sec:app-rl}). Because the rest of the page serves as the reference, the fault can be judged without a target. The consistency test holds \RJCtLocal\ such faults (icon \RJCtIcon, accent \RJCtColour, material \RJCtMaterial, radius \RJCtRadius).

\begin{figure*}[!tp]
\centering
\includegraphics[width=\textwidth]{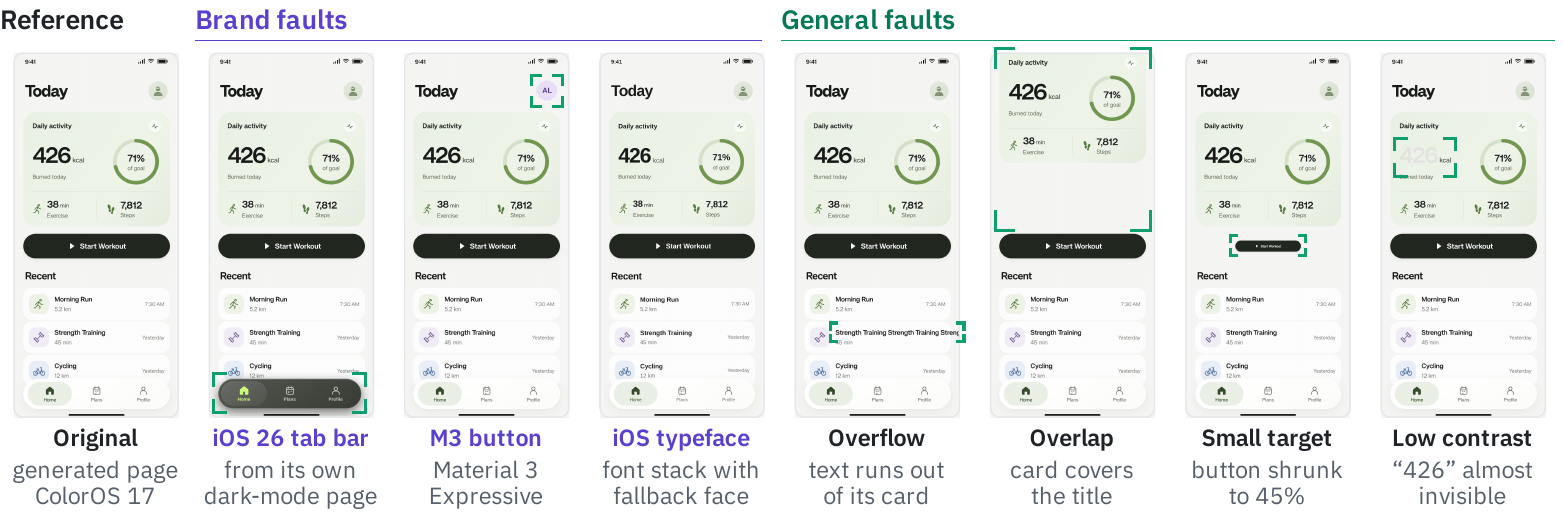}
\caption{Faults made by the data engine on one base page (a generated ColorOS 17 page); each copy changes one thing in the page code and carries an exact label (amber corner marks). A component can be transplanted, with all its styles, from a page of another design system (the iOS 26 tab bar keeps the dark glass of its own dark-mode page); general faults are wrong in any design.}
\label{fig:engine}
\end{figure*}

\paragraph{Transplanted components.}
One component is replaced by the same kind lifted from a page of another design system (\cref{fig:engine}), with all computed styles inlined and its \texttt{::before}/\texttt{::after} content materialized so it renders as on its source page; its post-layout box is the label. Nothing is pasted onto pixels, so no compositing trace can act as a cue. The consistency test holds \RJCtSwap\ transplanted components.

\begin{figure*}[t]
\centering
\includegraphics[width=\textwidth]{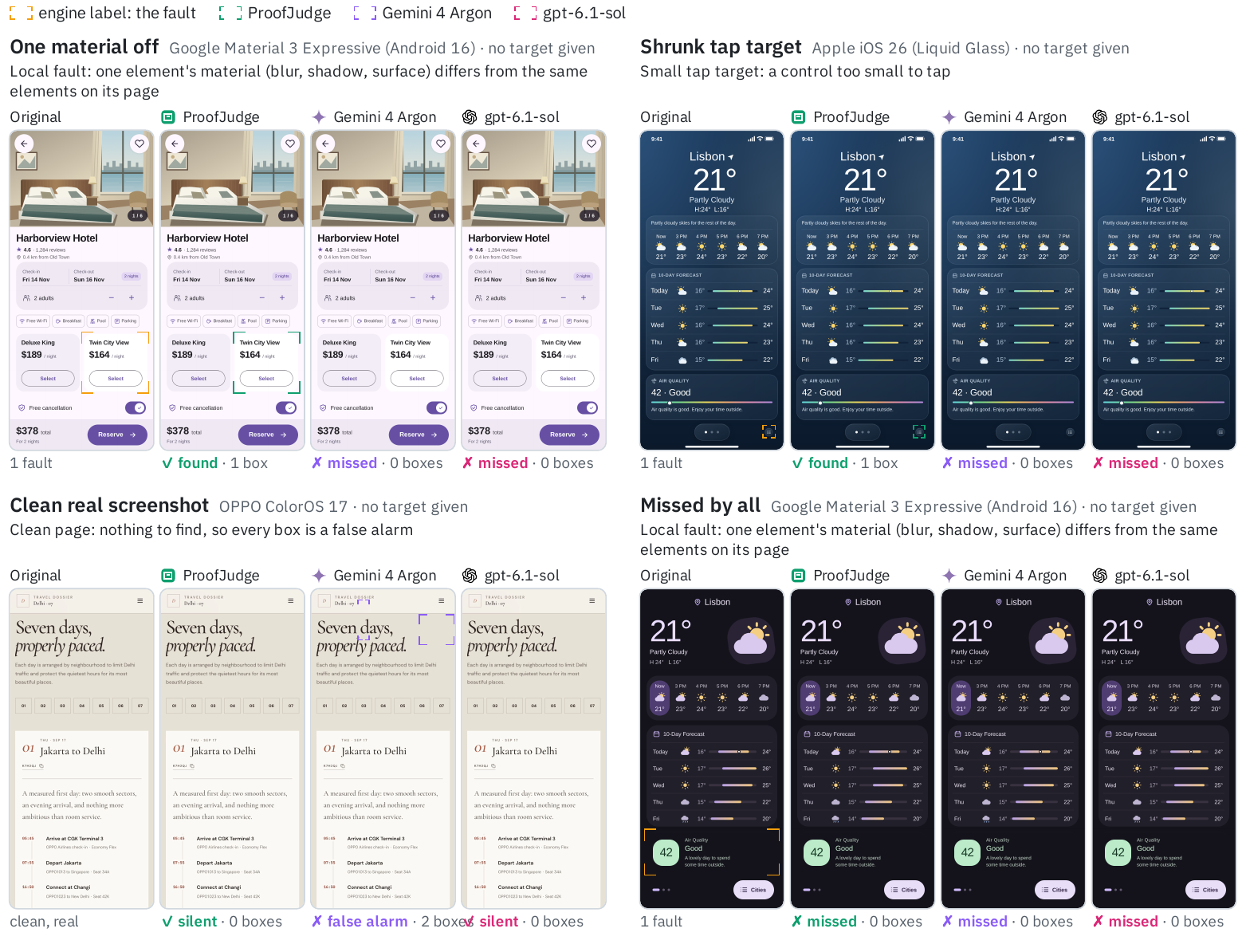}
\caption{Where \judge\ succeeds and fails on the consistency test (no target given): four hand-picked cases, three of its strengths and one fault every judge misses; \cref{tab:consistency} gives the overall picture, where \judge\ has the highest F1 but Gemini 4 Argon finds more local style faults. Under each title, one line names the fault; each example shows the original screenshot with the engine's exact label (amber corner marks), then \judge\ and two API judges with only their own boxes. \judge\ alone finds the room card whose surface differs from its sibling's and a control shrunk below tap size (IoU $\geq$ 0.5), whereas the API judges' boxes, if any, lie elsewhere; on a clean real screenshot it stays silent where Gemini 4 Argon boxes a timeline and an empty area. No judge finds the air-quality card that has lost the surface its sibling cards keep. Twenty cases chosen by a fixed rule, including failures, are in \cref{sec:app-qual}.}
\label{fig:qual}
\end{figure*}

\paragraph{Text layout faults.}
Each fault changes the CSS of one text element of a held-out base page; the label is the box of that element's text after layout.
Seven rules are wrong on their own: the characters of a label touch, the lines of a paragraph touch, a short control label wraps, one word or character stands alone on the last line, a word is broken inside, Chinese punctuation starts a line, and a label is cut with an ellipsis although its row has room.
Two rules are wrong only against the siblings: one of at least three like labels gets wider letter spacing, or one of at least three paragraphs a larger line height; for these, a hard negative applies the same change to every sibling, so the page stays consistent (\cref{sec:app-text-layout}).

\paragraph{\method-Train.}
The engine also makes a released training set: \XTrRows\ rows on \XTrImages\ images of \XTrPages\ pages (\XTrGenPages\ generated, \XTrRealPages\ public web pages), \XTrClean\ clean, with \XTrPairs\ designed differences each paired with a faulted twin (\cref{tab:train}; \cref{sec:app-train}); it trains the paper's \judge.

\section{What Judges Find}
\label{sec:findings}

\begin{figure*}[!tp]
\centering
\includegraphics[width=\textwidth]{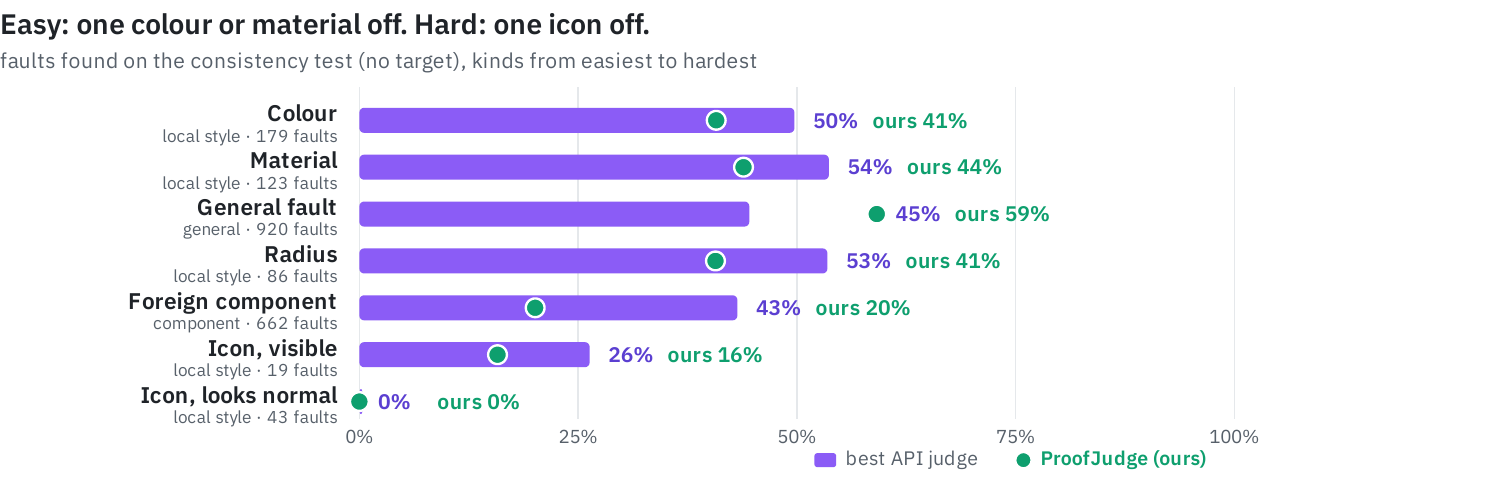}
\caption{What is easy and what is hard (consistency test, \RJPages\ screenshots, no target). Each row is one kind of fault, ordered by the mean share the four API judges find, easiest first; the bar is the share the best of them finds (Gemini 4 Argon, gpt-6.1-sol, Gemini 3.8 Flash, GPT-6 Luna), the dot the share \judge\ finds. One element with another colour, material or corner radius is found by the best API judge about half the time; a foreign icon is hard for every judge, and none of the four nor \judge\ finds an icon that looks normal on its page (Claude Opus 5.5, which flags most clean pages, finds \XOpusIconBrandHit\ of \XIconBrandN; \cref{tab:fault-recall}). \judge\ finds the most general faults (\RJGenRec). Every judge's numbers are in \cref{tab:fault-recall}, the split of icon faults in \cref{sec:app-visibility}.}
\label{fig:fault-recall}
\end{figure*}

\begin{figure}[!tbp]
\centering
\includegraphics[width=\columnwidth]{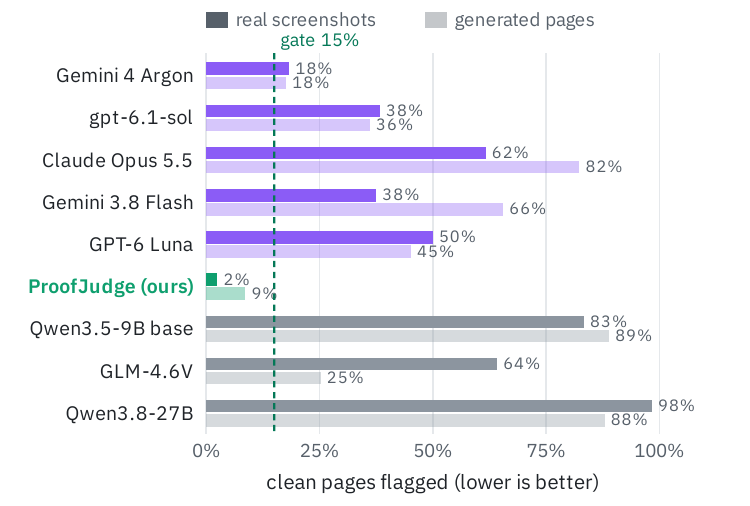}
\caption{Clean pages flagged on the consistency test (no target; \RJCleanPages\ clean pages: \RJCleanReal\ real, \RJCleanGen\ generated), for each judge one bar for clean real screenshots (dark) and one for clean generated pages (light). The dashed line is the 15\% gate; among judges that find faults, only \judge\ stays within it on both.}
\label{fig:fa}
\end{figure}

We test five frontier models from three companies, nine open VLMs and the open base Qwen3.5-9B on the target-free tests, all with the same prompt.

\begin{table*}[t]
\centering\tstyle
\caption{Consistency test (\RJPages\ screenshots, \RJCleanPages\ of them clean: \RJCleanReal\ real and \RJCleanGen\ generated; no target; same prompt; local models greedy without thinking, API judges at default reasoning). Consistency recall: swapped components and local style faults. The gate: at most 15\% of real and of generated clean pages flagged; only \judge\ with the cut passes it among fault-finding judges. No answer: unparsable, scored as finding nothing. Cut: a finding is kept only when the verdict's probability of a clean page is below a threshold chosen on validation. Seconds per page: \judge\ (rank-16 adapter) on one A100-80GB, open VLMs on A100-40GB, Gemini via API. Tinted: over half of the clean pages flagged.}\label{tab:consistency}
\begin{tabular}{@{}lcccccccccc@{}}
\toprule
 & \multicolumn{2}{c}{Clean pages flagged} & \multicolumn{2}{c}{Recall} & & & & & \multicolumn{2}{c}{Speed} \\
\cmidrule(lr){2-3}\cmidrule(lr){4-5}\cmidrule(l){10-11}
Judge & real $\downarrow$ & generated $\downarrow$ & general $\uparrow$ & consistency $\uparrow$ & Precision $\uparrow$ & F1 $\uparrow$ & F1$_{.3}$ $\uparrow$ & No answer $\downarrow$ & s/page & GPUs \\
\midrule
\tgroup{11}{API judges}
\mGA & 18\% & 18\% & 38\% & 41\% & 35\% & 0.37 & 0.43 & 0 & 6.1 & -- \\
\mSOL & 38\% & 36\% & 45\% & 32\% & 25\% & 0.30 & 0.36 & 3 & -- & -- \\
\mCL\ (medium) & \lo{62\%} & \lo{82\%} & 49\% & \textbf{49\%} & 22\% & 0.31 & 0.35 & 0 & -- & -- \\
\mGF & 38\% & \lo{66\%} & 40\% & 26\% & 22\% & 0.26 & 0.29 & 0 & 20.6 & -- \\
GPT-6 Luna & 50\% & 45\% & 31\% & 17\% & 20\% & 0.22 & 0.27 & 1 & -- & -- \\
\tgroup{11}{\judge\ (ours, Qwen3.5-9B + LoRA)}
\mRJ{(ours)} & \textbf{2\%} & \textbf{9\%} & \textbf{59\%} & 25\% & \textbf{62\%} & \textbf{0.48} & \textbf{0.50} & 0 & 1.0 & 1 \\
\mQW & \lo{83\%} & \lo{89\%} & 21\% & 33\% & 2\% & 0.04 & 0.05 & 282 & -- & -- \\
\tgroup{11}{Open VLMs, zero-shot}
\mlogo{qwen}Qwen3.8-27B & \lo{98\%} & \lo{88\%} & 29\% & 20\% & 11\% & 0.15 & 0.17 & 0 & 4.83 & 2 \\
InternVL3.5-38B & \lo{95\%} & \lo{99\%} & 19\% & 17\% & 3\% & 0.05 & 0.06 & 16 & 4.93 & 4 \\
GLM-4.6V (106B MoE) & \lo{64\%} & 25\% & 14\% & 6\% & 5\% & 0.07 & 0.08 & 10 & 0.71 & 8 \\
InternVL3.5-8B & \lo{93\%} & \lo{94\%} & 13\% & 27\% & 2\% & 0.04 & 0.05 & 182 & 7.45 & 1 \\
Mistral-Small-3.2-24B & \lo{99\%} & \lo{100\%} & 2\% & 4\% & 0\% & 0.01 & 0.02 & 7 & 6.55 & 2 \\
\mlogo{qwen}Qwen3-VL-32B & \lo{79\%} & 36\% & 26\% & 28\% & 3\% & 0.05 & 0.06 & 199 & 8.64 & 4 \\
\mlogo{qwen}Qwen3-VL-30B-A3B & 27\% & 28\% & 8\% & 8\% & 2\% & 0.03 & 0.04 & 629 & 4.24 & 2 \\
\mlogo{qwen}Qwen3-VL-8B & 23\% & 0\% & 2\% & 0\% & 3\% & 0.02 & 0.02 & 14 & 1.67 & 1 \\
UI-TARS-1.5-7B (GUI agent) & 8\% & 0\% & 0\% & 0\% & 0\% & 0.00 & 0.00 & 4 & 0.76 & 1 \\
\bottomrule
\end{tabular}
\end{table*}

\paragraph{Knowing is not finding.}
Asked in words, frontier models answer every brand-free design question they parse (28--31 of 31), and 41--59 of 70 when they must name the defect in a red box on a screenshot; training lifts Qwen3.5-9B from 33 to 60 of 70 on the latter (\judge\ v5). These probes suggest a gap between knowing what a defect is and finding it on an unmarked page without false alarms.

\paragraph{Critics find much, and much that is not there.}
Frontier judges flag \XApiRealFA\ of clean real screenshots and \XApiGenFA\ of clean generated pages (\cref{tab:consistency}; \cref{fig:fa}), and four of the nine open VLMs flag nearly every clean page.
Gemini 4 Argon comes closest to the gate (\XArgonRealFA\ and \XArgonGenFA); only our trained \judge\ passes it among fault-finding judges (\cref{sec:rolejudge}). Are the labels right? One of us checked \RJHumanChecked\ test items by hand, the engine's change against its page and clean pages for anything wrong, and found \RJHumanDrop\ mislabelled: a clean page with a fault, or a change that is hidden or still looks fine. With them left out, every F1 changes by at most 0.01 (\judge\ \RJHumanFone, Gemini 4 Argon \RJHumanArgonFone) and the order of the judges stays. Most false alarms are thus the judges', though some label noise remains.

\paragraph{One element against its own page.}
Without a target (\cref{fig:fault-recall}), the two strongest API judges find a changed corner radius (Gemini 4 Argon \XArgonRadius, gpt-6.1-sol \XSolRadius) at least as often as a general fault (\XArgonGenRec, \XSolGenRec), but rarely a foreign icon (\XArgonIcon, \XSolIcon). Many local faults are subtle: a colour fault changes a thin accent inside a large control (a median 2.6\% of the box's pixels, $\Delta E$ 1.0 over the box), and many icon faults look normal on their page, foreign only by a design language's conventions (\cref{sec:app-visibility}).

\begin{table*}[t]
\centering\tstyle
\caption{Text layout test (held-out page requirements, no target, IoU $\ge 0.5$): recall per rule, and pages flagged among 75 clean pages and 71 hard negatives (every sibling changed alike). 367 faults: characters touch 66, letter spacing unlike siblings 46, lines touch 20, line height unlike siblings 17, short label wraps 23, lone last word or character 73, word broken inside 50, needless ellipsis 72. Spacing: characters touch, letter spacing wider than the siblings'; Lines: lines touch, line height larger than the siblings'; Breaks: a short label wraps, a lone last word or character, a word broken inside, a needless ellipsis; Flagged: clean pages and hard negatives flagged. The base pages are English, so the Chinese line-start punctuation rule has no test item; ``--'': not run.}
\label{tab:text-layout}
\begin{tabular}{@{}lccccccccccc@{}}
\toprule
Judge & \multicolumn{2}{c}{Spacing $\uparrow$} & \multicolumn{2}{c}{Lines $\uparrow$} & \multicolumn{4}{c}{Breaks $\uparrow$} & All $\uparrow$ & \multicolumn{2}{c}{Flagged $\downarrow$} \\
\cmidrule(lr){2-3}\cmidrule(lr){4-5}\cmidrule(lr){6-9}\cmidrule(l){11-12}
 & touch & wide & touch & tall & wrap & lone & word & ellipsis & & clean & neg. \\
\midrule
\mGA & 17\% & 41\% & 60\% & 18\% & 43\% & 22\% & 86\% & 81\% & 47\% & 32\% & 56\% \\
\mSOL & 27\% & 59\% & 70\% & 35\% & 26\% & 25\% & 100\% & 81\% & 54\% & 49\% & 68\% \\
\mCL\ (medium) & 73\% & 80\% & 70\% & 35\% & 43\% & 55\% & 98\% & 82\% & 72\% & 80\% & 96\% \\
\mRJ{(ours)} & 5\% & 46\% & 5\% & 0\% & 26\% & 14\% & 94\% & 96\% & 43\% & 15\% & 13\% \\
\bottomrule
\end{tabular}
\end{table*}

\paragraph{Text layout.}
On the text layout test (\cref{tab:text-layout}), Gemini 4 Argon finds \TArgonAll\ and gpt-6.1-sol \TSolAll\ of the faults: nearly every mid-word break and needless ellipsis, but few labels with touching characters, while flagging \TArgonNegFA\ and \TSolNegFA\ of the hard negatives, where all siblings share the same wide spacing. Both find most broken words (\TArgonWord, \TSolWord) but few lone last words (\TArgonLone, \TSolLone) and few labels whose characters touch (\TArgonTouch, \TSolTouch); at the weakest strengths they find almost none (characters touch at $-0.08$\,em: \TArgonTouchWeak, \TSolTouchWeak). Claude Opus 5.5 at medium effort finds the most (\TOpusAll, characters touching \TOpusTouch) but boxes almost every page: \TOpusCleanFA\ of the clean pages and \TOpusNegFA\ of the hard negatives. \judge\ finds \TRJAll: more needless ellipses than any API judge (\TRJEllipsis) and most broken words (\TRJWord), but few labels whose characters touch (\TRJTouch); it flags \TRJCleanFA\ of the clean pages and \TRJNegFA\ of the hard negatives, fewer than any API judge. Scores at IoU $\ge 0.3$ change little (Argon \TArgonAllThirty, Sol \TSolAllThirty).

\paragraph{Against a named brand.}
Naming a target brand makes the task harder (\cref{sec:app-brand}): models know a design language about as well as its public documentation states it (\XKnowMaterial\ on Material 3 Expressive, \XKnowMagic\ on MagicOS 11), and frontier critics with a target flag 54--89\% of clean pages.
\section{\judge: a Small Consistency Check}
\label{sec:rolejudge}

\paragraph{Model and training.}
\judge\ is Qwen3.5-9B (thinking disabled) with a LoRA of rank 128 on all linear language-model layers and a frozen vision encoder (\cref{fig:rolejudge}).
It outputs JSON with a verdict and findings, each giving a box in 0--1000 screen coordinates, the element, the fault family and a reason, requiring no detection head.
Our main model receives no target. It is trained by SFT on the \XTrRows\ rows of \method-Train (\XTrClean\ clean; \cref{tab:train}), with one to four independent faults per page (general, display, local style and text layout); no test or validation page requirement is in its data (\cref{sec:app-v7}).
Two runs that differ only in the seed land up to 0.04 F1 apart on validation, so we train twice and average the two adapters (a weight average of their updates), each at its final step.
A finding is kept only when the model's probability that the page is clean, read from its verdict, is below \RJCut; the cut was chosen on held-out validation screens, never on the test (tested variants: \cref{sec:app-v7}). Earlier runs with a brand target (v5) or one fault per page (v6) are in \cref{sec:app-v5,sec:app-rl}.

\begin{table}[!tbp]
\centering\tstyle\setlength{\tabcolsep}{4pt}
\caption{Capacity and seeds (v7f data, two epochs, each run at its final step 2,570; LoRA $\alpha$ = 2 $\times$ rank). Validation: F1 with each model's own cut on the validation split. Pair: the average of the two seeds' adapters. Test: the 2,765-page consistency test with the validation cut. Rank 64 was tested and rank 256 validated once on the training machine; their raw answers are not released. $^\dagger$The rank-16 model tested is one run, step 2,500, checkpoint and cut chosen on validation (the paper's model before this one).}\label{tab:capacity}
\begin{tabular}{@{}lccccc@{}}
\toprule
LoRA rank & seed 42 & seed 7 & pair & test F1 & \shortstack{95\% interval\\vs rank 16} \\
\midrule
16 & 0.556 & 0.517 & 0.550 & 0.405$^\dagger$ & -- \\
64 & 0.561 & 0.550 & 0.571 & 0.445 & $[+0.025, +0.055]$ \\
128 (ours) & 0.586 & 0.598 & \textbf{0.607} & \textbf{0.481} & $[+0.060, +0.091]$ \\
256 & 0.605 & 0.576 & 0.599 & -- & -- \\
\bottomrule
\end{tabular}
\end{table}

\paragraph{A consistency check.}
A 9B model does not learn eight design languages from public pages, but it can compare an element against its own page, so \judge\ asks only which elements are wrong on the page itself.
On the consistency test (\cref{tab:consistency,fig:qual}) it reaches F1 \RJFone, the highest of all judges, ahead of Gemini 4 Argon (\XArgonFone; difference \RJBootDiff, 95\% interval $[\RJBootLo, \RJBootHi]$ over \RJBootN\ paired page resamples), Claude Opus 5.5 (\XOpusFone) and gpt-6.1-sol (\XSolFone), and far above its 9B base (\XBaseFone) and the open VLMs (\XOpenFone).
It is the only fault-finding judge within the 15\% gate, flagging \RJRealFA\ of clean real screenshots and \RJGenFA\ of clean generated pages (Argon: \XArgonRealFA\ and \XArgonGenFA), with the highest precision (\RJPrec; Argon \XArgonPrec).
It finds the most general faults (\RJGenRec; API judges \XApiGenRec) but fewer local style faults than the strongest API judges (\RJConRec; Argon \XArgonConRec), and it stays ahead at IoU $\geq 0.3$.
The cut keeps it quiet on clean pages: without it the same model flags \RJRawRealFA\ and \RJRawGenFA\ of clean pages, yet its F1 (\RJRawFone) is still above every other judge, so the lead does not come from the cut alone. Capacity does the rest: the same data with a rank-16 LoRA (one run, its checkpoint and cut chosen on validation) flags \RJPrevRealFA\ and \RJPrevGenFA\ with the cut and reaches F1 \RJPrevFone; F1 rises from rank 16 to 128 and stops there (\cref{tab:capacity}).

\begin{table}[t]
\centering\tstyle\setlength{\tabcolsep}{4pt}
\caption{External benchmark (consistency prompt unchanged). OwlEye: 600 Android screenshots, half with display issues; flagging all scores F1 0.67. Full table: \cref{tab:owleye-full}.}\label{tab:owleye}
\resizebox{\columnwidth}{!}{\begin{tabular}{@{}lccc@{}}
\toprule
 & \multicolumn{3}{c}{OwlEye} \\
\cmidrule(l){2-4}
Judge & F1 $\uparrow$ & Normal flagged $\downarrow$ & Accuracy $\uparrow$ \\
\midrule
\tgroup{4}{API judges}
\mGA & \textbf{0.86} & 28\% & \textbf{84\%} \\
\mGF & 0.75 & \lo{64\%} & 68\% \\
\mSOL & 0.76 & \lo{63\%} & 68\% \\
\mCL\ (medium) & 0.79 & \lo{53\%} & 73\% \\
\tgroup{4}{\judge\ (ours)}
\mRJ{(ours)} & 0.72 & \textbf{25\%} & 73\% \\
\mQW & 0.64 & \lo{97\%} & 48\% \\
\tgroup{4}{Open VLMs, zero-shot}
GLM-4.6V & 0.74 & \lo{66\%} & 66\% \\
\mlogo{qwen}Qwen3-VL-8B & 0.73 & 33\% & 72\% \\
\mlogo{qwen}Qwen3.8-27B & 0.66 & \lo{100\%} & 50\% \\
\multicolumn{4}{@{}l}{\quad six other open VLMs (7B--38B): F1 0.05--0.66} \\
\bottomrule
\end{tabular}}\end{table}

\paragraph{External benchmark.}
On OwlEye's real Android screenshots~\cite{owleye} (\cref{tab:owleye}), where flagging all scores F1 0.67, \judge\ reaches F1 \XOwFone\ and flags \XOwFA\ of normal screenshots (accuracy \XOwAcc), fewer than any API or open judge with recall above 50\%; Gemini 4 Argon leads (F1 0.86).

\paragraph{Do engine-made faults predict real ones?}
Over all judges scored on both (Spearman $\rho$), a judge's rank on the consistency test predicts its rank on OwlEye's real screenshots: $\rho=\XTrOwRho$ between the two F1 scores over \XTrOwN\ judges and variants, and $\rho=\XTrFaRho$ between the shares of clean real screenshots and of OwlEye's normal screens flagged. Engine-made faults rank critics as real screens do on whether a page is clean.

\begin{figure*}[!tp]
\centering
\includegraphics[width=\textwidth]{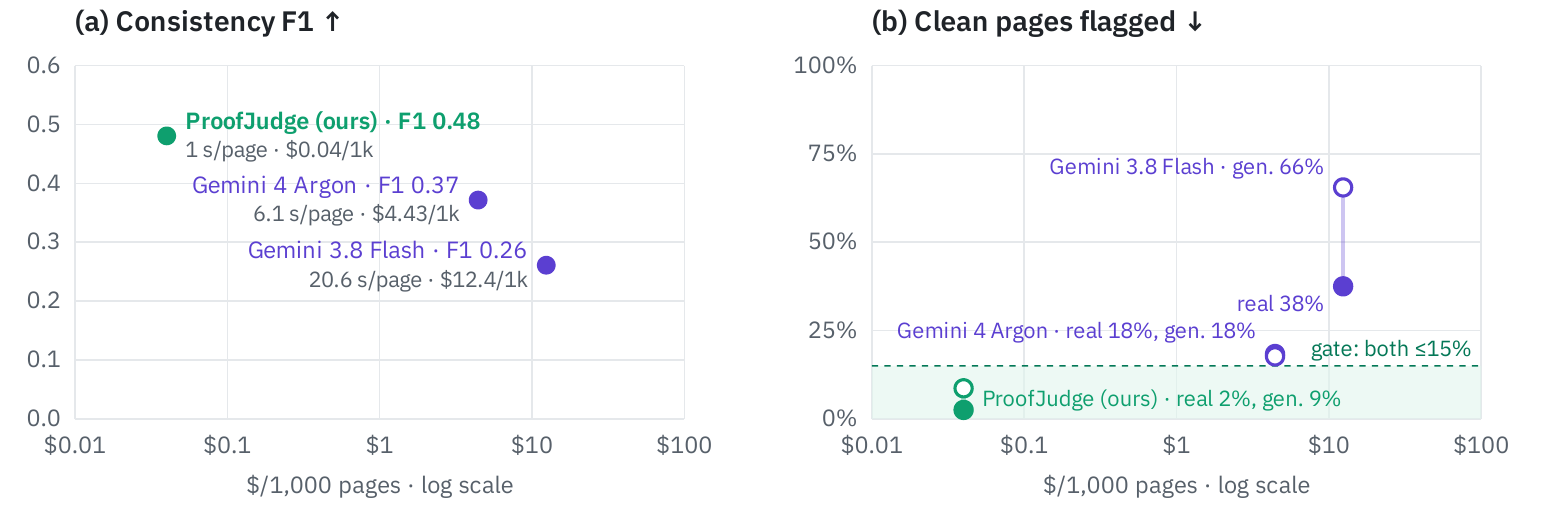}
\caption{Cost against accuracy, the same cost scale in both panels. \textbf{(a)} F1 on the consistency test (\RJPages\ pages, no target given), with time per page and cost per 1,000 pages. \textbf{(b)} Clean pages flagged: real screenshots (filled) and generated pages (open); a checkpoint is usable only below the dashed line on both. \judge\ is the cheapest by two orders of magnitude and flags the fewest clean pages, real and generated. \judge's time was measured on v6, with the same base model and a smaller adapter (rank 16). \judge's cost assumes \$2 per A100-hour; the API costs are list prices on 2026-10-25.}
\label{fig:cost}
\end{figure*}

\paragraph{Cost and speed.}
On one A100-80GB, \judge\ (timed with an earlier rank-16 adapter) takes \XRJSec\,s per page and, at 16 concurrent requests and an assumed \$2 per GPU-hour, costs about \$\XRJCost\ per 1,000 pages, versus \XArgonSec\,s and \$\XArgonCost\ for Gemini 4 Argon (\cref{fig:cost}), while keeping private files in house.

\begin{figure}[t]
\centering
\includegraphics[width=\columnwidth]{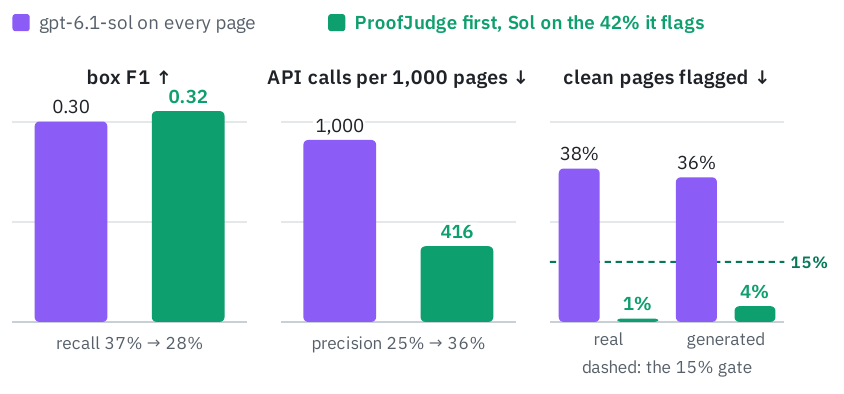}
\caption{\judge\ as a cheap first check (consistency test, no target). gpt-6.1-sol runs either on every page or only on the \XSolGateSent\ of pages \judge\ flags: box F1 \XSolAloneFone\ to \XSolGateFone, precision \XSolAlonePrec\ to \XSolGatePrec, recall \XSolAloneRec\ to \XSolGateRec, and clean pages flagged \XSolAloneRealFA\ to \XSolGateRealFA\ (real) and \XSolAloneGenFA\ to \XSolGateGenFA\ (generated; dashed line: the 15\% gate). Sol reports no token counts, so the saving is shown in API calls; the same gate with Gemini 4 Argon is in the text.}
\label{fig:gate}
\end{figure}

\paragraph{A cheap first check.}
In front of an API judge, \judge\ removes most false alarms (\cref{fig:gate}; consistency test).
When gpt-6.1-sol looks only at the \XSolGateSent\ of pages that \judge\ flags, box F1 rises from \XSolAloneFone\ to \XSolGateFone\ and precision from \XSolAlonePrec\ to \XSolGatePrec, while recall falls from \XSolAloneRec\ to \XSolGateRec; clean pages flagged fall from \XSolAloneRealFA\ to \XSolGateRealFA\ of real screenshots and from \XSolAloneGenFA\ to \XSolGateGenFA\ of generated pages, with \XGateCalls\ instead of 1,000 API calls per 1,000 pages.
Behind the same gate Gemini 4 Argon flags \XGateRealFA\ and \XGateGenFA\ of clean pages at \$\XGateCost\ instead of \$\XArgonCost\ per 1,000 pages, but its F1 falls from \XArgonFone\ to \XGateFone, below \judge\ alone.
As a second opinion, \judge\ finds \XUnionOnly\ faults that Argon misses, most of them general: together they reach recall \XUnionRec\ (Argon \XArgonRec) and F1 \XUnionFone, above Argon but below \judge\ alone, and flag \XUnionRealFA\ of clean real screenshots and \XUnionGenFA\ of clean generated pages.

\section{Find, then Fix}
\label{sec:fix}

A critic is useful if its findings help a generator repair the page. We test this where a repair can be checked exactly (\cref{sec:app-fix}): \XFixFaultPages\ consistency-test pages with one local fault whose right value the page itself shows (one corner radius, accent or surface unlike its siblings), and \XFixCleanPages\ clean generated pages, on which a false finding should change nothing.

\paragraph{Fixer and hints.}
gpt-6.1-sol gets the page's HTML, its screenshot and one hint: none (fix whatever does not match the rest of the page, or return it unchanged), the true finding, or a judge's findings with the instruction to fix the reported problems that are real and change nothing else. As a \emph{gate}, a judge only decides whether a page goes to the fixer, which then works with no hint.
A fault is fixed when the pixels inside its box that differ from the clean page fall to a quarter or less; a page is harmed when more than 1\% of the pixels outside the box, or anywhere on a clean page, change.

\paragraph{Repair is easy; knowing when to repair is not.}
With the page's code and no hint, the fixer fixes \XFixNone\ of the faults and never harms the rest of a faulty page, but it also changes \XFixNoneCleanHarm\ of the clean pages. Only the true finding gives both a high fix rate (\XFixOracle) and no observed harm to the \XFixCleanPages\ clean pages (\XFixOracleCleanHarm).
A judge's findings help only as far as they are right: under ``change nothing else'' a finding on the wrong element keeps the fixer from the real fault, so Gemini 4 Argon's findings lower the fix rate to \XFixArgon. \judge\ is quiet on these subtle local faults (\cref{fig:fault-recall}): its findings let the fixer repair \XFixRJv\ and harm \XFixRJvCleanHarm\ of clean pages (Argon: \XFixArgonCleanHarm).
As a gate, Gemini 4 Argon sends \XFixGateArgonSent\ of the pages and \XFixGateArgon\ of the faults are fixed; \judge\ sends \XFixGateRJvSent, harms \XFixGateRJvClean\ of clean pages, and lets only \XFixGateRJv\ of the faults be fixed (\cref{sec:app-fix}).
The value of a critic for repair therefore lies both in staying silent on clean pages and in finding the small faults, and no judge does both yet.

\paragraph{Cost.}
Per 1,000 pages, fixing every page costs about \$\XFixCostNone, an Argon gate \$\XFixCostGateArgon\ and a \judge\ gate \$\XFixCostGateRJv\ (a fixer call about \$\XFixCallCost), paid for by the faults each gate misses; without code (a screenshot or an installed app) the judge's location matters more (untested).

\section{Limitations and Conclusion}
\label{sec:conclusion}

\paragraph{Limitations.}
We study single phone screens without interaction or motion. Clean labels rest on generated base pages that two judges found on-spec. Scores come from model judges, and most earlier \judge\ variants were trained once (the capacity runs and the paper's model use two seeds). The text layout test uses English pages only, and in the fix test the fixer reads the page's code, which a screenshot or an installed app does not offer.

\paragraph{Conclusion.}
No critic yet boxes every fault on a screen without a reference: critics flag many clean pages, an element that breaks its own page's style often escapes every judge, and text layout faults are found unevenly. A small trained critic has the highest F1 on the consistency test and is the only fault-finding judge quiet on clean pages, yet it misses many local style and text layout faults and lags on real screens. Its potential use is a cheap first pass over generated pages, before a human review: it passes most clean pages and points to the element to fix. Whether this saves designers time is untested; a study with designers is future work.

{
    \small
    \bibliographystyle{ieeenat_fullname}
    \bibliography{main}
}

\section*{Supplementary Material}   %
\appendix
\renewcommand{\thefigure}{S\arabic{figure}}\setcounter{figure}{0}
\renewcommand{\thetable}{S\arabic{table}}\setcounter{table}{0}
\section{A short primer on the tasks}
\label{sec:app-primer}
This section explains the benchmark for readers outside UI design; the main paper gives the formal definitions (\cref{sec:task}).

\paragraph{Design languages.}
Every phone maker publishes, or keeps in house, a \emph{design language}: the rules that make its apps look like one family. They fix how controls look (a switch, a tab bar, a search field), the accent colour and how it is used, the corner radius of cards and buttons, the material of surfaces (blur, glass, shadow), the icon set and the type scale. The rules change between versions, so a target is always a brand \emph{and} a version, for example iOS 26 (Liquid Glass) or ColorOS 17. A screen that ignores them still works, but looks foreign, like an Android switch inside an iOS app. We cover eight languages: Google's Material 3 Expressive and the systems of seven phone makers (\cref{fig:dataset}).

\paragraph{Three kinds of errors.}
Three families of faults need no target. A \emph{general} fault is wrong under any design language: text that runs off the screen, two elements that overlap, a button too small to tap, text too faint to read. A \emph{local style} fault is visible without knowing any brand: one element differs from the elements of the same kind on its own page, for example one button with sharp corners on a page of round ones, or a component comes from another design system. A \emph{text layout} fault breaks a rule of type setting, such as a word broken inside or a needless ellipsis (\cref{tab:taxonomy}). A \emph{brand} fault needs a named target and is only in \cref{sec:app-brand}.

\paragraph{Task 1, knowledge: does the model know the rules?}
Without any screenshot, the model answers multiple-choice questions written from the official guidelines, such as \emph{which brand prescribes this corner radius for cards?} (one of eight) or \emph{which of these four statements is true for MagicOS 11?} (one of four). Picture questions then ask the same on screenshots: which brand's rule a screenshot shows, which brand a red-boxed element comes from, and which design dimension of a page breaks a given target. A model that cannot answer these cannot be expected to judge a page.

\paragraph{Task 2, critique: can the model point at what is wrong?}
The model gets one screenshot and no target. For every problem it draws a box around the element, names its family (general, local style or text layout) and says what is wrong. This is what a design reviewer does. Because the faults are made by our data engine in the page's code, we know exactly which element is wrong and where, so the answer is scored automatically: a box counts as found when it is matched one to one to a true element with intersection over union $\geq 0.5$ ($\geq 0.3$ as a lenient variant, \cref{sec:app-iou}). \emph{Recall} is the share of true faults found, \emph{precision} the share of drawn boxes that are right, F1 combines the two, and a \emph{false alarm} is a ``faulty'' verdict or any box on a clean page, which has nothing to find. A useful critic needs both: a model that boxes everything finds every fault and is useless to a designer.

\paragraph{Task 3, design: can the model build to the rules?}
The model sees a real screenshot of a brand's app and rebuilds it as a web page, with only the brand name or also with a short public summary of the brand's rules (a \emph{spec card}). Two judge models score the rebuild item by item: is every element of the original there, in the right place and style (likeness), and does the page follow the card (specification), quoting the card for every violation.

\paragraph{The data engine and the quality gate.}
Labelling problems on real screenshots by hand is slow and uncertain. We instead start from good pages and break exactly one thing in their code: transplant a component from a page of another design system, change every card's radius, make one title overflow, or give one button another radius than its siblings (\cref{fig:engine}). The change is the label. Before a page is used, two judges check that it follows its brand's card; pages that fail this \emph{quality gate} are dropped, because a page that is already off-brand cannot be a clean example.

\paragraph{Data.}
All data (pages we generate and screenshots collected from the web) are released.

\paragraph{\judge.}
\judge\ is our small baseline critic: a 9-billion-parameter open vision-language model (Qwen3.5-9B) that we fine-tune on engine copies to answer the critique task in the same box format (\cref{fig:rolejudge}). It is meant to be cheap enough to run on every page in-house; the main paper reports where it is and is not good enough. The paper's model is v7f, trained by SFT without a target (\cref{sec:app-v7}).

\section{Positioning against prior work}
\label{sec:app-related}

\Cref{tab:related} compares the works discussed in the related work of the main paper, feature by feature.

\paragraph{UI understanding and visual grounding.}
UI datasets such as Rico~\cite{deka2017rico} and VINS~\cite{bunian2021vins} provide screens and component boxes, and screen models such as ScreenAI~\cite{baechler2024screenai} and Ferret-UI~\cite{you2024ferretui} understand and ground UI elements. GUI agents ground instructions to screen locations~\cite{hong2024cogagent,cheng2024seeclick,gou2025uground}, building on grounded MLLMs~\cite{peng2023kosmos2}. They locate an element that matches a description; we locate elements that are wrong on their own page or under any design, without a reference, and name the fault.

\section{Dataset}
\label{sec:app-data}

\begin{figure*}[!tp]
\centering
\includegraphics[width=\textwidth]{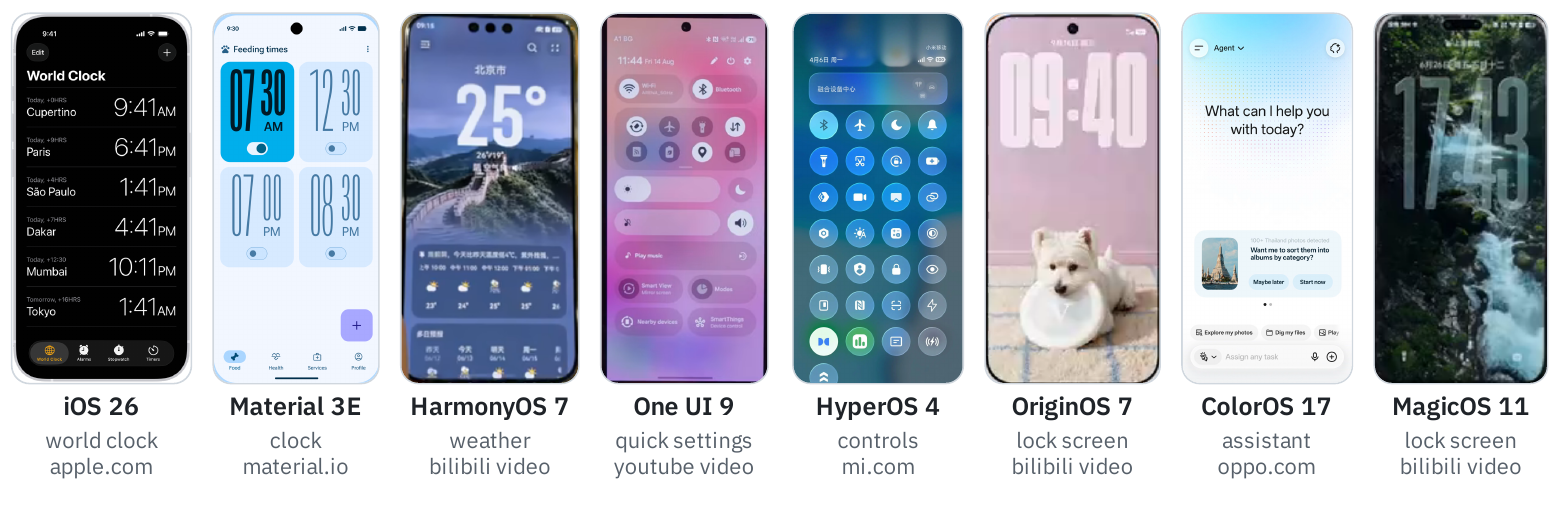}
\caption{One accepted web screenshot per design language at the benchmark's target version. Each label gives the version, the screen and its source; frames from videos are marked. Screens keep the language they were published in.}
\label{fig:dataset}
\end{figure*}

\begin{figure}[!tbp]
\centering
\includegraphics[width=\columnwidth]{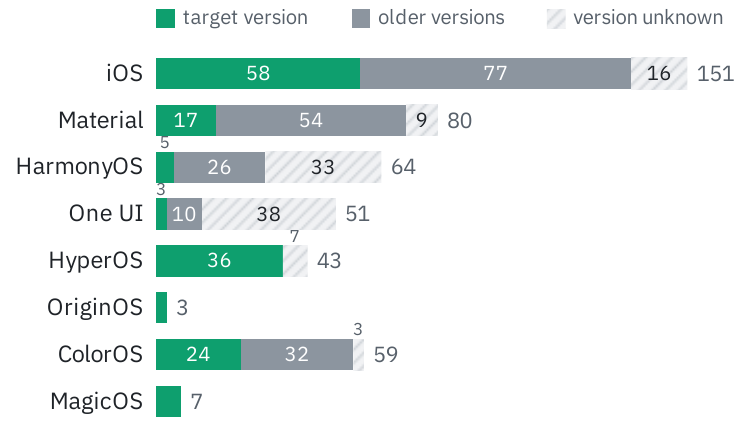}
\caption{Accepted web screenshots per design language (snapshot of 23 October 2026): the target version, older versions, and screenshots whose page names no version. 458 in all.}
\label{fig:datasetcounts}
\end{figure}

\paragraph{Public part: brand screenshots.}
We searched each brand's own sites for phone screenshots: product and release pages, user guides, newsroom images and design-guideline sites.
Every image found is recorded in a manifest with its source page, date of collection, terms of use, size and hash.
An image is kept only if it is a single portrait phone screen; composites of several screens or devices, landscape images and images whose content is not a phone UI are set aside.
Each kept screenshot is labelled with the system version that its page title or URL names (for example a version-pinned iPhone User Guide URL); when the page names no version, the screenshot is kept with version unknown and is not used where a version is required.
In the snapshot of 23 October 2026 (from a frozen manifest copy identified by its SHA-256; collection continues) the manifest held 8,755 images from brand sites, design guides, open-source repositories and videos, of which 458 were accepted, 352 with a named version (\cref{fig:datasetcounts}); 4,980 were composites, 1,852 were not portrait, 1,325 were rejected and 140 were still pending at the snapshot. Most accepted screenshots come from official sites (320) and design guides or developer documents (81); 24 are video frames, all of vivo's and Honor's among them.
\Cref{fig:dataset} shows one accepted screenshot per brand at the benchmark's target version.

\paragraph{Public part: other sources.}
Screens from public UI datasets are recorded in the same manifest with their licences (3,327 files at the time of writing, mostly from iOS-1K, Design2Code and Sketch2Code); spec cards summarise each version's official design guidelines in seven fixed fields (shape, colour, material, navigation, controls, type, spacing), each fact with the URL of the official page it comes from and nothing filled in from memory or press; a field the official pages do not state is marked \emph{not specified} (4 such items for iOS 26, 1 for Material 3 Expressive, 8 for ColorOS 17, whose only public specification covers home-screen cards); the other five cards are written the same way, and since vivo, Xiaomi and Honor publish only partial official documents, most of their fields are \emph{not specified}; pages built with open-source UI libraries (\eg Framework7) and their fault copies are generated by us.

\paragraph{Release.}
\method\ is released in full, images included, with the source, date and terms of use of each item.

\begin{figure*}[!tp]
\centering
\includegraphics[width=\textwidth]{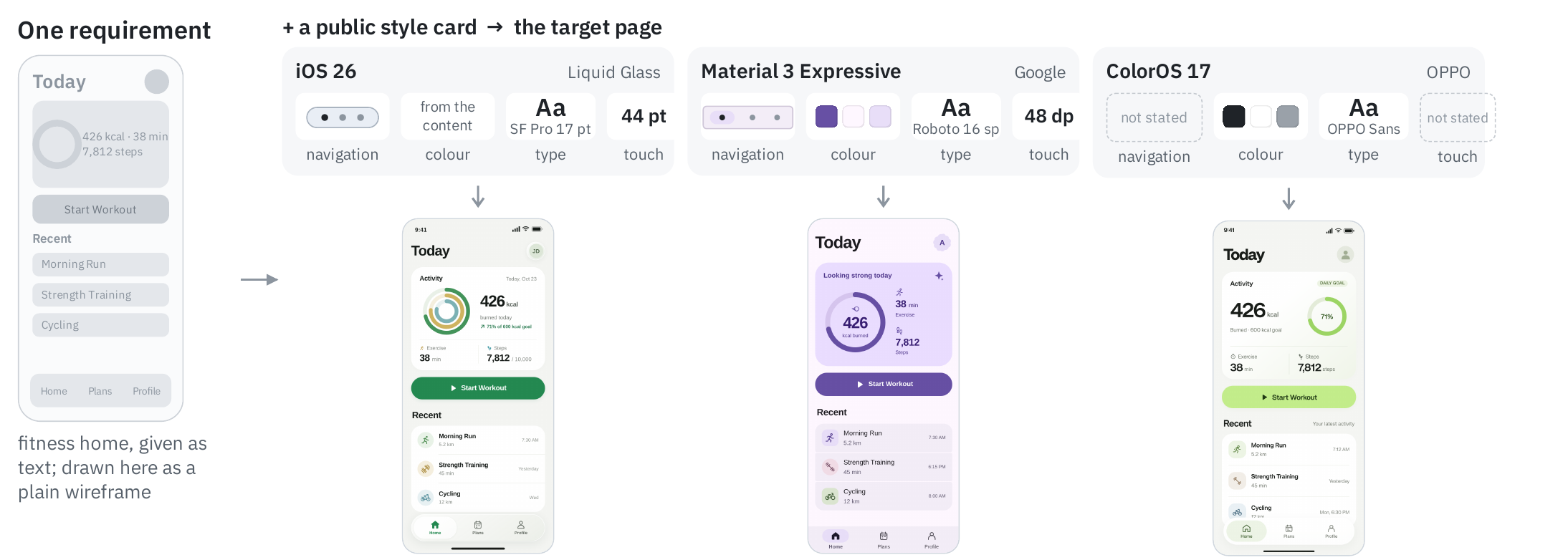}
\caption{One requirement (text, drawn as a wireframe) plus a public style card gives a target page in that design language (generated pages). The ColorOS 17 card cannot state the navigation shape or selected state, as OPPO's public pages do not, and the page still draws a selected pill.}
\label{fig:brands}
\end{figure*}

\section{Brand conformance}
\label{sec:app-brand}
The main paper asks the target-free question. This section reports the brand tracks of \method: a judge or generator is given a named target $T=(b,v)$, a brand and its version, \eg iOS 26 Liquid Glass or ColorOS 17, and a brand error is wrong only with respect to that target.
A useful brand review must say which convention of which design language an element breaks; the split into general and brand errors yields a testable property: when the same screenshot is judged against a different brand, general findings should stay and brand findings should change.

\paragraph{Three tasks.}
\emph{Knowledge} asks what a model knows of each design language, without a page to judge: it \emph{states} conventions (text-only multiple choice built from public spec cards, brand words removed), \emph{recognises} them on real screenshots (name the brand and version; find the odd screen out of four), and \emph{applies} them to images (which brand's rule a screenshot shows; which brand a red-boxed swapped element follows; which design dimension of a page breaks a target). A general probe asks the same of brand-free design knowledge, in words and on a red-boxed defect.
\emph{Critique} is the localization task of the main paper with a named target added, on engine copies with exact labels.
\emph{Design} asks a generator to restore a real screenshot of a brand's current version as HTML, given the brand name only or the public card; two judges score likeness to the original (content, layout and style, checked against an element inventory) and to the specification (card fields followed, each violation quoting the card).
Every task reports general and brand errors separately. The test sets hold 1,316 text questions from 658 facts of the official guidelines, 680 picture-and-text items and 101 general-design items for knowledge; 643 screenshots (147 clean) for critique against a named brand and \RJPages\ (\RJCleanPages\ clean; the first version of this test had 576 pages) for the consistency question; and 47 real screens for design.

\paragraph{Targets.}
A brand target always specifies a brand \emph{and} version, because a brand's versions differ in materials, corner radii, navigation and accent use; \cref{fig:brands} shows how much a single page changes across targets.
The benchmark covers seven main brands: Apple iOS, Huawei HarmonyOS, OPPO ColorOS, vivo OriginOS, Xiaomi HyperOS, Samsung One UI and Honor MagicOS, each in its latest released version (iOS 26, HarmonyOS 7, ColorOS 17, OriginOS 7, HyperOS 4, One UI 9 and MagicOS 11), alongside Material 3 Expressive.

\paragraph{Kinds of errors.}
Main \Cref{tab:taxonomy} splits target-free faults into general, local style and text layout faults; brand faults, which need a named target, are reported only here.
\emph{General UI errors} are wrong under any design language and are labelled independently of the brand.
\emph{Brand errors} are wrong only with respect to the target brand and version, and fall into six common visual design dimensions: component, colour, radius, material, icon and type; spacing is a general error.
Layout (alignment, spacing, margins, density, tap targets) is general knowledge: its rules are shared across brands.

\begin{figure}[!tbp]
\centering
\includegraphics[width=\columnwidth]{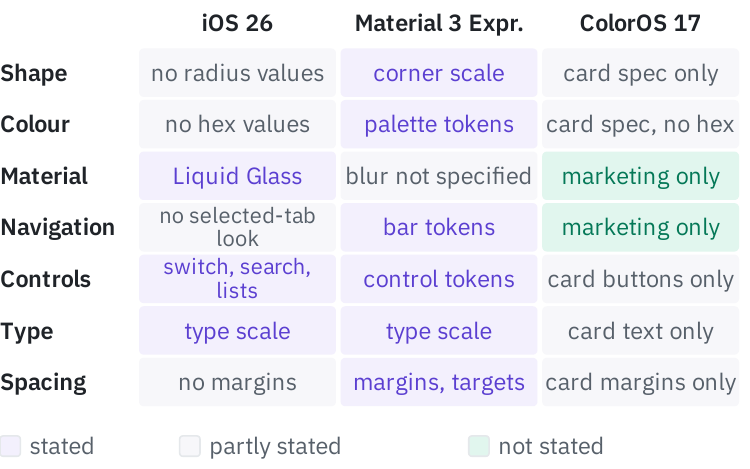}
\caption{How much each brand's official public pages specify per field of our spec cards. Partly stated: the card marks part of the field as unspecified. For ColorOS 17, the only public specification covers home-screen cards; material and navigation appear only in marketing text.}
\label{fig:specgap}
\end{figure}
\begin{table*}[!tp]
\centering\tstyle
\caption{Knowledge probes. \emph{State}: 1,316 text questions from 658 facts of the official guidelines, half asking which brand a fact belongs to (8-way) and half which of four facts is the brand's; the 124-question card pilot is in \cref{sec:app-pilot5}. \emph{Apply} v2: 680 items on test-split screenshots. \emph{Apply}: which brand's rule a real screenshot shows (no brand given), which brand a red-boxed swapped element follows, and which design dimension of a page breaks a given target. \emph{General}: brand-free design knowledge in words (say) and on a red-boxed defect (see); ``--'': not run.}\label{tab:knowledge}
\begin{tabular}{@{}lcccccc@{}}
\toprule
 & State & \multicolumn{3}{c}{Apply (picture + card text)} & \multicolumn{2}{c}{General design} \\
\cmidrule(lr){2-2}\cmidrule(lr){3-5}\cmidrule(l){6-7}
Model & which brand / fact $\uparrow$ & whose rule $\uparrow$ & boxed source $\uparrow$ & broken dimension $\uparrow$ & say $\uparrow$ & see $\uparrow$ \\
\midrule
Chance & 19\% & 13\% & 13\% & 20\% & 25\% & 17\% \\
\midrule
\mSOL & \textbf{1000}/1316 & 229/412 & 33/163 & \textbf{37}/105 & 28/28 & 46/69 \\
\mGPT & 983/1316 & 242/412 & 39/163 & 33/105 & 31/31 & 48/70 \\
\mGA & 925/1315 & \textbf{285}/408 & \textbf{53}/163 & 22/105 & 31/31 & 59/70 \\
\mGP & 816/1316 & 209/408 & 25/163 & 25/105 & 31/31 & 41/70 \\
\mGF & 755/1316 & 221/408 & 32/163 & 25/105 & 31/31 & 50/70 \\
\midrule
\mQW & 303/1316 & 77/410 & 18/162 & 22/105 & 28/31 & 33/70 \\
\mRJ{v5, earlier, brand target} & 295/1316 & 117/412 & 17/163 & 27/105 & 27/31 & \textbf{60}/70 \\
\bottomrule
\end{tabular}
\end{table*}

\paragraph{Documentation.}
The knowledge gap follows public guidelines: official pages leave 1 spec-card item unstated for Material 3 Expressive, 4 for iOS 26, and 8 for ColorOS 17 (\cref{fig:specgap}).
Given only the name ``OPPO ColorOS 17'', a frontier generator drew the bottom navigation of ColorOS 16 and Material, and the three reviewing judges missed the difference too.

\paragraph{Quality gate.}
A base page is a positive only if it follows its card: two judges check it item by item, as in the design task, and admit only pages with a specification score of at least 6.0.
80\% of the 223 checked pages pass (one Material 3 Expressive dark page, the file browser, was not generated): all iOS 26 and Material 3 Expressive pages, 93\% of ColorOS 17 and One UI 9, 68\% of HarmonyOS 7, 64\% of OriginOS 7 and HyperOS 4, and 57\% of MagicOS 11, matching the order of the knowledge results.
The gate scores only what the card states and cannot vouch for undocumented conventions, so weakly documented brands also have the most uncertain positives.

\paragraph{Type swaps (brand, type).}
The page's font family and heading weights are replaced with another brand's.
Because proprietary fonts cannot be redistributed (\eg SF Pro), public-part type faults use only open fonts or real screenshots.

\paragraph{Style dimensions (brand: colour, radius, material, icon).}
One visual dimension is measured on another brand's page for the same requirement and applied to the target page: the accent of coloured controls, the corner radius of each component kind, the blur, shadow and background of cards and bars, or the icons in corresponding positions. Each modified element becomes a labelled box, and any box with unchanged pixels is dropped. These page-wide faults leave the page internally consistent, making them purely brand questions. The brand test holds 352 such boxes (accent 129, icon 103, material 91, radius 29) and 52 with a page-wide typeface.

\paragraph{Frontier generation is clean but off-brand.}
Looking only for general defects, two independent API judges find one on 2 of 24 first-round generated pages (8\%, 0.08 per page) and on 7--9 of 64 generated base pages (11--14\%, 0.16--0.23 per page, mostly overlap); \judge\ v5 finds none on the former.
Rebuilds recover 96--98\% of a screen's text.
Brand styling is the gap: 20\% of generated base pages fail the specification gate, and 32--43\% for HarmonyOS 7, HyperOS 4, OriginOS 7 and MagicOS 11; three judges identify the intended brand on only 34\% of them (iOS 89\%, Material 100\%, the others 6--35\%); and rebuilds carry about two off-spec elements per page (\cref{tab:restore}).
Frontier generators make general UI faults on a minority of pages; conformance to a specific design language is the larger gap.

\begin{figure}[!tbp]
\centering
\includegraphics[width=\columnwidth]{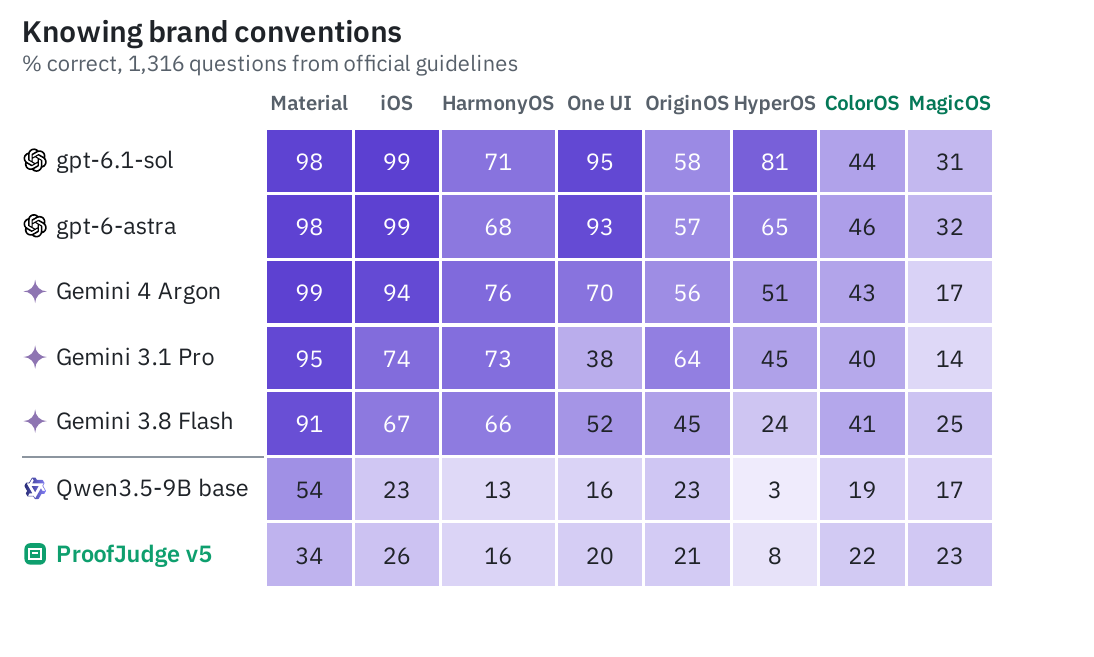}
\caption{Knowing brand conventions, per design language: \% correct on 1,316 text questions from 658 facts of the official guidelines. Frontier models reach 91--99\% on Material 3 Expressive but 40--46\% on ColorOS 17 and 14--32\% on MagicOS 11 (green: the two least documented); the open 9B models, the base and \judge\ v5, stay near chance. Brands at their latest versions: iOS 26, Material 3 Expressive, HarmonyOS 7, One UI 9, OriginOS 7, HyperOS 4, ColorOS 17, MagicOS 11.}
\label{fig:knowledge-heat}
\end{figure}

\paragraph{Knowledge follows public documentation.}
On 1,316 text questions built from 658 facts in the official guidelines (\cref{tab:knowledge}; per brand in \cref{fig:knowledge-heat}), every frontier model answers 91--99\% of the Material 3 Expressive questions and 67--99\% of the iOS 26 ones, but only 40--46\% for ColorOS 17 and 14--32\% for MagicOS 11; asked which brand a MagicOS fact belongs to, models are right 0--10\% of the time.
Even the number of facts the guidelines yield follows documentation, from 298 questions for iOS to 74 for HyperOS 4.
Knowing a rule is not applying it to pixels (680 picture-and-text items on test-split screenshots): given the target and a page with one broken dimension, models pick the broken dimension on only 21--35\% of 105 items (chance 20\%), and name whose rule a ColorOS 17 screenshot shows on 9--27\% of 96, against 94--100\% for iOS.
Recency matters too: the Gemini judges always chose a brand's oldest version (\cref{sec:app-pilot5}).
The open 9B models stay near chance: Qwen3.5-9B answers 23\% and \judge\ v5 (earlier, brand target) 22\%, against 76\% for gpt-6.1-sol, and LoRA training runs stay within one point of their base.

\paragraph{General design knowledge is high in words.}
Every probed frontier model answers each brand-free text question it parses (28--31 of 31), and 41--59 of 70 when naming the defect in a red box; training lifts Qwen3.5-9B from 33 to 60 of 70 on the latter (\judge\ v5), while its brand knowledge stays flat.

\paragraph{Critique against a named brand.}
On defect test v2 (\cref{tab:defect}), frontier models recall \XDtGenRec\ of general faults and \XDtBrandRec\ of brand faults, yet every model raises false alarms on over half the clean pages (54--89\%), drawing 2 to 6 boxes per page.
Precision therefore stays at \XDtPrec, and box F1 at \XDtFone.

\paragraph{Design: likeness separates, the specification does not.}
Rebuilding 47 real screens of six brands as HTML, the GPT models place and style almost every element (layout 98--99\%, style 96--98\%) while Gemini 3.1 Pro and 3.8 Flash misplace about 30\% and Gemini 4 Argon about 10\%, yet frontier rebuilds score about as high on the specification as the real screens (6.9--7.7 against 7.4; \cref{sec:app-restore}).

\section{Design task: restore scores}
\label{sec:app-restore}
\begin{table*}[!tp]
\centering\tstyle\setlength{\tabcolsep}{6pt}
\caption{Restore track (47 screens): itemised likeness and specification scores (1--10), mean of two judges (Gemini 3.1 Pro, gpt-5.6-sol); Viol.\ is the number of card violations per page. Tinted: below 75\%.}\label{tab:restore}
\begin{tabular}{@{}llrcccccc@{}}
\toprule
Generator & Ctx & $n$ & Content & Layout & Style & Likeness & Spec. & Viol. \\
\midrule
\mGPT & card & 47 & 100\% & 98\% & 98\% & \textbf{9.9} & 7.5 & 1.8 \\
 & name & 47 & 100\% & 99\% & 96\% & 9.8 & 7.3 & 2.0 \\
\mSOL & card & 47 & 100\% & 99\% & 97\% & \textbf{9.9} & 7.4 & 1.8 \\
 & name & 47 & 100\% & 99\% & 97\% & 9.9 & 7.3 & 1.9 \\
\mGA & card & 47 & 98\% & 89\% & 95\% & 9.4 & 7.4 & 1.9 \\
 & name & 47 & 98\% & 90\% & 95\% & 9.4 & 7.3 & 2.0 \\
\mGF & card & 47 & 99\% & \lo{74\%} & 87\% & 8.7 & 7.5 & 1.6 \\
 & name & 46 & 99\% & \lo{72\%} & 86\% & 8.6 & 7.4 & 2.0 \\
\mGP & card & 47 & 97\% & \lo{69\%} & 88\% & 8.5 & \textbf{7.7} & 1.8 \\
 & name & 47 & 96\% & \lo{69\%} & 88\% & 8.4 & 6.9 & 2.1 \\
\mQW & card & 38 & 81\% & \lo{46\%} & \lo{43\%} & 5.7 & 7.0 & 2.2 \\
 & name & 36 & \lo{54\%} & \lo{23\%} & \lo{22\%} & 3.3 & 5.4 & 2.4 \\
\midrule
Real screenshots & -- & 47 & -- & -- & -- & -- & 7.4 & 1.7 \\
\bottomrule
\end{tabular}
\end{table*}

In the design task a generator rebuilds a real screenshot of a brand's current version as HTML, with the brand name only or with its public spec card (47 screens of six brands; OriginOS 7 and MagicOS 11 publish no app-page screenshots).
Two judges score every page item by item (\cref{tab:restore}): likeness checks content, layout and style against an inventory of the original screen, and the specification score counts card fields followed, every violation quoting the card verbatim.
The GPT models place and style almost every element (layout 98--99\%, style 96--98\%), Gemini 3.1 Pro and 3.8 Flash misplace about 30\% (layout 69--74\%) and Gemini 4 Argon about 10\%, and the open base loses most of the screen.
Frontier rebuilds score about as high on the specification as the real screens (7.3--7.7 against 7.4; Gemini 3.1 Pro with the name only 6.9), so that score is no ceiling and does not separate them; the card raises it for Gemini 3.1 Pro (6.9 to 7.7) and the open base (5.4 to 7.0) but hardly for the GPT models.
Qwen3.5-9B did not finish 20 of its 94 rebuilds; its rows average the finished ones.

\paragraph{Judges differ in strictness.}
Scored as whole pages rather than item by item, the real screenshots of the design task average 8.0 under Gemini 3.1 Pro and 5.4 under gpt-5.6-sol on the same 1--10 scale, and the scorers of an earlier generation pilot agreed only moderately (rank correlation 0.47--0.81), so all models are scored by the same judges.

\section{Knowledge: recency or documentation?}
\label{sec:app-pilot5}

On real screenshots, judges named the brand among eight (chance 1/8) and, given the brand, the version among three (1/3); a few unparsed answers lower some denominators.

\begin{table*}[!tp]
\centering\tstyle
\caption{Brand and version recognition on real screenshots. ColorOS 15 has only 9 screenshots. Tinted: below half.}\label{tab:pilot5}
\begin{tabular}{@{}lcccccc@{}}
\toprule
 & iOS 18 & iOS 26 & iOS 27 & ColorOS 15 & ColorOS 16 & ColorOS 17 \\
\midrule
\tgroup{7}{Brand among eight}
\mGF & 30/30 & 30/30 & 30/30 & \lo{4/9} & 13/16 & \lo{5/24} \\
\mGP & 30/30 & 30/30 & 30/30 & \lo{3/9} & \lo{5/16} & \lo{10/24} \\
\mGPT & 30/30 & 30/30 & 30/30 & \lo{1/9} & \lo{7/16} & \lo{5/24} \\
\mQW & 29/29 & 25/25 & 30/30 & \lo{1/9} & \lo{0/15} & \lo{2/23} \\
\mRJ{v1 (earlier)} & 30/30 & 29/30 & 28/30 & \lo{1/9} & \lo{1/16} & \lo{1/24} \\
\midrule
\tgroup{7}{Version, brand given}
\mGF & 30/30 & \lo{0/30} & \lo{0/30} & 9/9 & \lo{0/16} & \lo{0/24} \\
\mGP & 30/30 & \lo{0/30} & \lo{0/30} & 9/9 & \lo{0/16} & \lo{0/24} \\
\mGPT & 30/30 & 27/30 & \lo{0/30} & \lo{2/9} & \lo{6/16} & \lo{1/24} \\
\mQW & 30/30 & \lo{0/30} & \lo{0/30} & \lo{4/9} & \lo{1/16} & \lo{5/24} \\
\mRJ{v1 (earlier)} & 30/30 & \lo{0/30} & \lo{0/30} & 8/9 & \lo{3/15} & \lo{2/24} \\
\bottomrule
\end{tabular}
\end{table*}

ColorOS is hard to recognise even in versions released before the judges' cutoffs, which points to thin documentation; the Gemini judges always answer the oldest version and gpt-6-astra (cutoff April 2026) names iOS 26 but not iOS 27, so recency matters too.

\begin{figure*}[!tp]
\centering
\includegraphics[width=\textwidth]{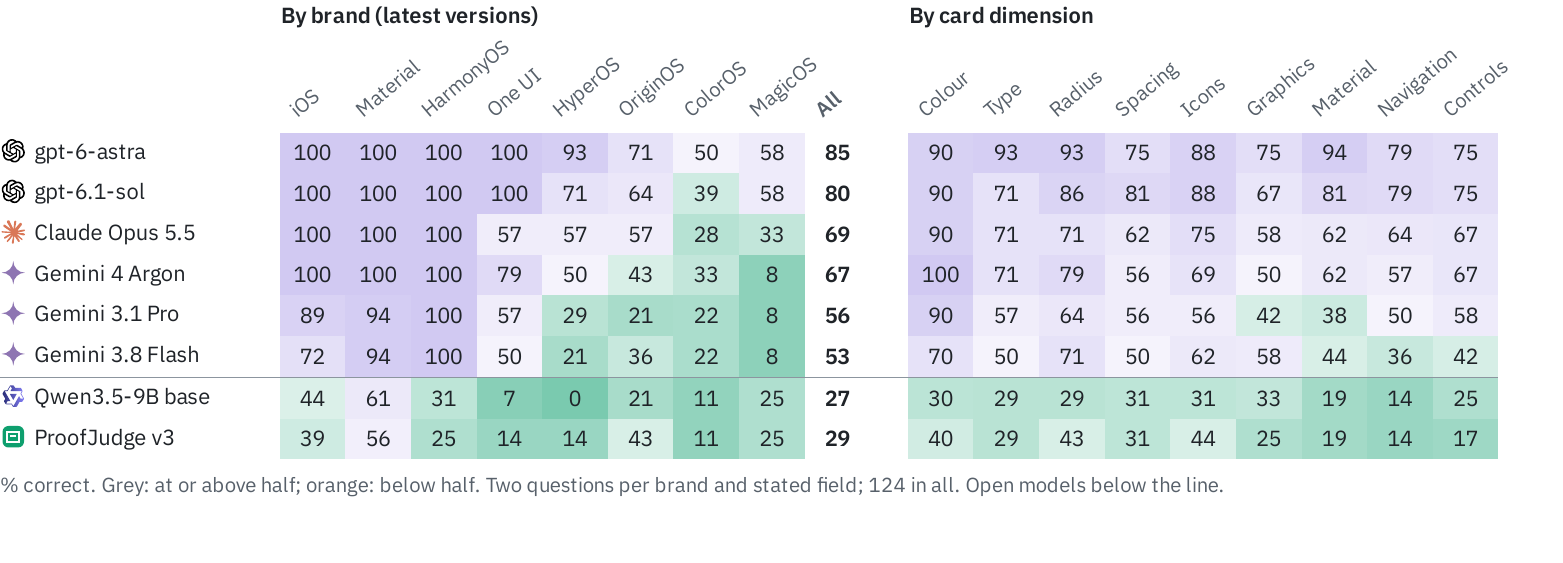}
\caption{\emph{State} probe (text only): accuracy of every model by brand and by card dimension, two questions per brand and stated card field, 12--18 per brand and 124 in all. Each brand at its latest version: iOS 26, Material 3 Expressive, HarmonyOS 7, One UI 9, HyperOS 4, OriginOS 7, ColorOS 17, MagicOS 11. Small and preliminary.}
\label{fig:knowledge}
\end{figure*}

\paragraph{Knowledge probe (state).}
Without images, multiple-choice questions built from the public cards (two per brand and stated field, 124 in all) were answered perfectly for HarmonyOS 7 (16/16 for all six frontier judges) and nearly so for Material 3 Expressive (17--18/18), but only 4/18 (Gemini 3.8 Flash), 4/18 (Gemini 3.1 Pro), 9/18 (gpt-6-astra), 7/18 (gpt-6.1-sol), 5/18 (Claude Opus 5.5) and 6/18 (Gemini 4 Argon) for ColorOS 17, and 1/12, 1/12, 7/12, 7/12, 4/12 and 1/12 for MagicOS 11. The open base Qwen3.5-9B answers 33 of the 124 (27\%).

\section{Name only vs.\ public card}
\label{sec:app-card}

Round-0 pages of an English generation pilot (four page requirements: fitness home, new event, gallery multi-select, music player) were generated twice, from the brand name only and from the name plus the public spec card (12 pages each: 4 requirements $\times$ 3 brands), and each set was scored twice by Gemini 3.1 Pro and Gemini 3.8 Flash (mean), once with the name only and once with the card in the judge's prompt.
Only 4 pages per brand: a preliminary sign, not a result.

\begin{table*}[!tp]
\centering\tstyle
\caption{Brand scores (1--10) of round-0 pages by generation context and judge context.}\label{tab:card}
\begin{tabular}{@{}lccc@{}}
\toprule
Judge context & Name-only pages & Card pages & Difference \\
\midrule
Name only & 5.88 & 6.29 & $+0.41$ \\
Name + card & 6.17 & 7.00 & $+0.83$ \\
\midrule
\tgroup{4}{By brand, name-only pages $\rightarrow$ card pages (name-only judge / card-aware judge)}
iOS 26 & \multicolumn{3}{c}{6.25 $\rightarrow$ 5.75 \quad / \quad 6.88 $\rightarrow$ 7.50} \\
Material 3 Expressive & \multicolumn{3}{c}{7.25 $\rightarrow$ 8.00 \quad / \quad 7.38 $\rightarrow$ 7.88} \\
ColorOS 17 & \multicolumn{3}{c}{4.12 $\rightarrow$ 5.12 \quad / \quad 4.25 $\rightarrow$ 5.62} \\
\bottomrule
\end{tabular}
\end{table*}

Under both judge contexts the card pages score higher, so the pages improve and not only the standard; the mixed difference (name-only pages scored without the card, card pages scored with it) would be about 1.1. ColorOS 17 gains most, in line with \cref{fig:specgap}. Judges without the card see half the gain, and for iOS 26 they prefer the name-only pages.

\section{OwlEye: all judges}
\label{sec:app-owleye}
\Cref{tab:owleye-full} lists every judge on OwlEye, including the six open VLMs summarised in one row of the main table.

\begin{table}[!tbp]
\centering\tstyle\setlength{\tabcolsep}{1.8pt}
\caption{OwlEye, all judges (\cref{tab:owleye}): 600 real Android screenshots, 300 with display issues (null values, missing images, occlusion, overlapping text, blur) and 300 normal; screenshot-level verdicts with the consistency prompt. FA: normal screenshots flagged. Images used for evaluation only. Flagging every screenshot scores F1 0.67 and accuracy 50\% on this balanced set, so read F1 with FA and accuracy; unanswered screenshots count as not flagged.}\label{tab:owleye-full}
\begin{tabular}{@{}lcccccc@{}}
\toprule
Judge & Recall & FA & Prec. & F1 & Acc. & No ans. \\
\midrule
\tgroup{7}{API judges}
\mGA & 96\% & \textbf{28\%} & \textbf{77\%} & \textbf{0.86} & \textbf{84\%} & 0 \\
\mGF & 99\% & \lo{64\%} & 61\% & 0.75 & 68\% & 0 \\
\mSOL & \textbf{100\%} & \lo{63\%} & 61\% & 0.76 & 68\% & 0 \\
\mCL\ (medium) & 99\% & \lo{53\%} & 65\% & 0.79 & 73\% & 0 \\
\tgroup{7}{\judge\ (ours)}
\mRJ{(ours)} & 71\% & 25\% & 74\% & 0.72 & 73\% & 0 \\
\mRJ{v6 (b2-950)} & 72\% & 37\% & 66\% & 0.69 & 68\% & 0 \\
\mRJ{v6 (c-2229)} & 81\% & \lo{64\%} & 56\% & 0.66 & 59\% & 0 \\
\mRJ{v5 (final)} & 97\% & \lo{87\%} & 53\% & 0.68 & 55\% & 0 \\
\mQW & 93\% & \lo{97\%} & 49\% & 0.64 & 48\% & 30 \\
\tgroup{7}{Open VLMs, zero-shot}
GLM-4.6V & 98\% & \lo{66\%} & 60\% & 0.74 & 66\% & 1 \\
\mlogo{qwen}Qwen3-VL-8B & 76\% & 33\% & 70\% & 0.73 & 72\% & 46 \\
\mlogo{qwen}Qwen3.8-27B & 99\% & \lo{100\%} & 50\% & 0.66 & 50\% & 1 \\
\mlogo{qwen}Qwen3-VL-32B & 89\% & \lo{80\%} & 53\% & 0.66 & 55\% & 50 \\
Mistral-Small-3.2 & 98\% & \lo{99\%} & 50\% & 0.66 & 50\% & 10 \\
InternVL3.5-38B & 98\% & \lo{100\%} & 50\% & 0.66 & 49\% & 6 \\
InternVL3.5-8B & 93\% & \lo{93\%} & 50\% & 0.65 & 50\% & 41 \\
\mlogo{qwen}Qwen3-VL-30B-A3B & 52\% & \lo{57\%} & 48\% & 0.50 & 48\% & 207 \\
UI-TARS-1.5-7B & 3\% & 1\% & 67\% & 0.05 & 51\% & 5 \\
\bottomrule
\end{tabular}
\end{table}

\section{How visible the local test faults are}
\label{sec:app-visibility}

\begin{table*}[t]
\centering\tstyle\setlength{\tabcolsep}{5pt}
\caption{Faults found by kind on the consistency test, every judge (no target, IoU $\ge 0.5$); columns from easiest to hardest by the mean of the four API judges, as in \cref{fig:fault-recall}; the two icon columns cover the 62 icon faults labelled for visibility (all icons: \cref{sec:app-visibility}). Faults per kind: colour 179, material 123, general 920, radius 86, foreign comp. 662, icon, visible 19, icon, looks normal 43. $^*$The base model boxes about eight elements per page, so its recall is not comparable.}\label{tab:fault-recall}
\begin{tabular}{@{}lccccccc@{}}
\toprule
Judge & Colour & Material & General & Radius & Foreign comp. & Icon, visible & Icon, looks normal \\
\midrule
\mGA & 50\% & 45\% & 38\% & 53\% & 43\% & 26\% & 0\% \\
\mSOL & 40\% & 54\% & 45\% & 47\% & 30\% & 21\% & 0\% \\
\mGF & 46\% & 33\% & 40\% & 12\% & 24\% & 21\% & 0\% \\
GPT-6 Luna & 32\% & 27\% & 31\% & 21\% & 13\% & 5\% & 0\% \\
\mCL\ (medium) & 65\% & 71\% & 49\% & 66\% & 48\% & 32\% & 7\% \\
\mRJ{(ours)} & 41\% & 44\% & 59\% & 41\% & 20\% & 16\% & 0\% \\
\mRJ{rank 16 (earlier)} & 37\% & 30\% & 51\% & 24\% & 14\% & 11\% & 0\% \\
\mQW$^*$ & 48\% & 28\% & 21\% & 45\% & 32\% & 16\% & 44\% \\
\bottomrule
\end{tabular}
\end{table*}

We measured each consistency-test fault against its clean page: the box's share of the screen, the share of its pixels that visibly change, and the mean CIEDE2000 colour difference over the box (medians; the same measures on the training faults beside them).
\emph{Colour.} A test colour fault boxes a large control (2.6\% of the screen) in which only 2.6\% of the pixels change ($\Delta E$ 1.0 over the box): a thin accent inside the control. A training colour fault boxes a small element (0.75\%) that changes a third of its pixels ($\Delta E$ 5.8), so the two are different tasks.
\emph{Icons.} We labelled each of the 62 icon faults of the first 576 test pages (the full test has \RJCtIcon) from crops of the faulty page: \emph{visible} when a viewer of that page alone can see the icon does not match it (another stroke weight, fill or colour, a duplicate or wrong-role glyph, an empty placeholder), \emph{brand only} when the icon looks normal on the page and only the brand's conventions mark it as foreign (a back arrow for a chevron, near-identical search glyphs, a menu icon for a map icon). Three AI judges (a Claude agent, Codex and Gemini 4 Argon) answered blind, from the faulty page only, with no model answer and no label shown; \BControlsN\ unchanged icons were mixed in, and none of them was called wrong (\BControls\ of \BControlsN). By majority \XIconVisN\ are visible and \XIconBrandN\ brand only; the judges agree pairwise with Cohen's $\kappa$ \BKappa, and their majority matches a first label by one annotator on \BAgree\ of 62. Gemini 4 Argon is both a judge here and a scored model. Neither the four API judges of \cref{tab:icon-split} nor \judge\ finds a brand-only icon. A check by designers is future work.
\emph{Clean pages.} The same three judges boxed anything wrong on the \BPages\ clean generated pages of the first 576 test pages, blind. On \BFaulty\ pages two judges boxed the same place (IoU $\ge 0.5$), for example a floating button over list rows or a send button cut off at the screen edge; \BUncertain\ more rest only on a clipped last column of a sideways-scrolling row, a usual scroll hint, and stay clean. With the faulty pages removed, Gemini 4 Argon flags \BArgonGen\ of the clean generated pages, gpt-6.1-sol \BSolGen\ and \judge\ \BRJGen\ (\BRJReal\ of the clean real ones); F1 on these pages stays at \BRJFone\ for \judge\ and \BArgonFone\ for Gemini 4 Argon, and only \judge\ passes the gate.
\emph{Swaps.} A test button swap changes 28\% of its box (training: 65\%), and test segmented controls and chips are a fifth and two fifths of the training size; navigation-bar swaps are alike in training and test, yet \judge\ finds few of them.
The pictures are 720$\times$1600, within the model's pixel limit, so nothing is lost to resizing.

\begin{table}[!tbp]
\centering\tstyle
\caption{Icon faults found on the first 576 pages of the consistency test (62 labelled icon faults; no target, IoU $\ge 0.5$), split by whether the page shows them. On all \RJCtIcon\ icon faults of the full test, Gemini 4 Argon finds \XArgonIcon\ and gpt-6.1-sol \XSolIcon. $^*$The base model draws about eight boxes per page, so its recall is not comparable.}\label{tab:icon-split}
\begin{tabular}{@{}lccc@{}}
\toprule
Judge & All icons $\uparrow$ & Visible $\uparrow$ & Brand only $\uparrow$ \\
\midrule
\mGA & \textbf{8\%} & \textbf{26\%} & 0\% \\
\mSOL & 6\% & 21\% & 0\% \\
\mGF & 6\% & 21\% & 0\% \\
GPT-6 Luna & 2\% & 5\% & 0\% \\
\mRJ{(ours)} & 5\% & 16\% & 0\% \\
\mQW$^*$ & 35\% & 16\% & 44\% \\
\bottomrule
\end{tabular}
\end{table}

\section{Text layout test}
\label{sec:app-text-layout}

The text layout test asks the consistency question (no target) about faults that are wrong under any design language and need no brand knowledge. Each fault changes the CSS of one text element of a held-out base page; the label is the box of that element's text after layout, and a fault that changes no visible pixel or lies below the first screen is dropped. Seven rules are wrong on their own: the characters of a label touch (letter spacing $-0.08$ or $-0.12$\,em), the lines of a paragraph touch (line height 0.8 or 0.9 of the font size), a short control label wraps onto two lines, one word or character stands alone on the last line, a word is broken inside, Chinese punctuation starts a line, and a label is cut with an ellipsis although its row has room (to 50 or 65\% of its width). Two rules are wrong only against the siblings: one of at least three labels of the same size, weight and colour gets wider letter spacing ($+0.06$ to $+0.3$\,em), or one of at least three paragraphs a line height of 1.8 or 2.2. For these two, a hard negative applies the same change to every sibling, so the page stays consistent.

On \TFaults\ faults (\cref{tab:text-layout}), Gemini 4 Argon finds \TArgonAll\ and gpt-6.1-sol \TSolAll\ at IoU $\ge 0.5$ (\TArgonAllThirty\ and \TSolAllThirty\ at IoU $\ge 0.3$). Both find most broken words (\TArgonWord, \TSolWord) and needless ellipses (\TArgonEllipsis, \TSolEllipsis), but few lone last words (\TArgonLone, \TSolLone) and few labels whose characters touch (\TArgonTouch, \TSolTouch); at the weakest strengths they find almost none (characters touch at $-0.08$\,em: \TArgonTouchWeak, \TSolTouchWeak; wider spacing at $+0.06$\,em: \TArgonSpaceWeak, \TSolSpaceWeak). They flag \TArgonCleanFA\ and \TSolCleanFA\ of the clean pages and more of the hard negatives (\TArgonNegFA, \TSolNegFA): spacing that is wide but the same on every sibling is often called a fault. Claude Opus 5.5 at medium effort finds the most (\TOpusAll, characters touching \TOpusTouch), but it boxes almost every page: \TOpusCleanFA\ of the clean pages and \TOpusNegFA\ of the hard negatives. \judge\ finds \TRJAll: more needless ellipses than any API judge (\TRJEllipsis) and most broken words (\TRJWord), but few labels whose characters touch (\TRJTouch); it flags \TRJCleanFA\ of the clean pages and \TRJNegFA\ of the hard negatives, fewer than any API judge. The test uses English pages only; Chinese pages are future work.

\section{Two IoU thresholds}
\label{sec:app-iou}
A finding matches a labelled element when their boxes overlap with IoU $\ge 0.5$ (main rule); \cref{tab:iou} gives every localisation score at both IoU $\ge 0.5$ and IoU $\ge 0.3$. Neither rule counts a small box whose centre merely lies inside a large faulty element. The lower threshold raises every judge's score, mostly for general faults with small true boxes, and leaves the order of the judges almost unchanged.

\begin{table*}[!tp]
\centering\tstyle
\caption{Localisation at both IoU thresholds (same answers scored twice). Consistency: \RJPages\ screenshots, no target; defect test v2: 643 screenshots with a named brand. --: not run.}\label{tab:iou}
\begin{tabular}{@{}lcccc@{}}
\toprule
 & \multicolumn{2}{c}{Consistency F1 $\uparrow$} & \multicolumn{2}{c}{Defect v2 F1 $\uparrow$} \\
\cmidrule(lr){2-3}\cmidrule(l){4-5}
Judge & IoU $\ge$ .5 & IoU $\ge$ .3 & IoU $\ge$ .5 & IoU $\ge$ .3 \\
\midrule
\mGA & 0.37 & 0.43 & 0.17 & \textbf{0.21} \\
\mSOL & 0.30 & 0.36 & 0.10 & 0.12 \\
\mGF & 0.26 & 0.29 & 0.09 & 0.12 \\
\mRJ{(ours)} & \textbf{0.48} & \textbf{0.50} & -- & -- \\
\mRJ{v5 (final)} & -- & -- & \textbf{0.19} & 0.20 \\
\mQW & 0.04 & 0.05 & 0.06 & 0.08 \\
\mlogo{qwen}Qwen3.8-27B & 0.15 & 0.17 & -- & -- \\
GLM-4.6V & 0.07 & 0.08 & -- & -- \\
\bottomrule
\end{tabular}\end{table*}

\section{Thinking effort of the API judges}
\label{sec:app-effort}
Thinking effort is not a hidden factor in our API results (\cref{tab:effort}). On a fixed 138-page subset of the consistency test (20 clean real, 30 clean generated, 20 swaps, 48 local faults, 20 general faults), we ran gpt-6.1-sol at five effort levels and Gemini 4 Argon at three, and repeated the default level of each.
The two runs of the same setting already differ by \SEffortNoise\ F1, and within that noise no level clearly beats medium (GPT) or default (Gemini), while higher levels are 2--5$\times$ slower.
The one difference that holds in both runs is that swapped components are found more often at medium or default (20--50\%) than at any other level (5--10\%). Every judge in the paper therefore runs at medium or default.

\begin{table*}[!tp]
\centering\tstyle
\caption{Thinking effort on 138 pages of the first 576-page consistency test (no target given). FA: clean pages flagged; Swap and Local: recall. Medium and default come from the full run (no timing).}\label{tab:effort}
\begin{tabular}{@{}llcccccc@{}}
\toprule
Judge & Effort & FA real & FA gen. & Swap & Local & F1 & s/page \\
\midrule
\mGA & low & 10\% & 20\% & 5\% & 27\% & 0.24 & 4 \\
 & default & 15\% & 30\% & 50\% & 33\% & \textbf{0.33} & -- \\
 & default, rerun & 5\% & 30\% & 40\% & 25\% & 0.25 & 5 \\
 & high & 15\% & 30\% & 5\% & 29\% & 0.22 & 19 \\
\midrule
\mSOL & low & 40\% & 27\% & 10\% & 33\% & 0.24 & 14 \\
 & medium & 35\% & 43\% & 20\% & 29\% & 0.19 & -- \\
 & medium, rerun & 45\% & 50\% & 20\% & 35\% & 0.19 & 14 \\
 & high & 40\% & 17\% & 10\% & 35\% & 0.22 & 27 \\
 & xhigh & 30\% & 27\% & 10\% & 35\% & 0.19 & 53 \\
 & max & 35\% & 27\% & 10\% & 40\% & 0.22 & 70 \\
\bottomrule
\end{tabular}
\end{table*}

\section{\judge\ v5 against a named brand}
\label{sec:app-v5}
On defect test v2 (\cref{tab:defect}), \judge\ v5 scores box F1 \XVfiveDtFone\ (Gemini 4 Argon: \XArgonDtFone) with the highest precision (\SVfivePrec), drawing about one box per page; it recalls \SVfiveGenRec\ of the general faults but only \XVfiveDtBrand\ of the brand faults.
It still flags 65\% of the clean pages: 28\% of the real screenshots but all 75 generated ones (the supervised start: 15\% and 57\%). It also reports about one finding per page: on the 85 pages with two to six faults it finds \SVfiveMulti\ and never all (gpt-6.1-sol: \SSolMulti, all on \SSolMultiAll\ pages). v7 trains on pages with one to four faults (\cref{sec:app-v7}).
Two causes add up (\cref{sec:app-rl}): the counterfactual training rows label the same clean generated picture \emph{follows} 515 times and \emph{does not follow} 1,030 times under other targets, which a model without brand knowledge cannot tell apart, and the GRPO reward pays a guessed box whenever the policy's belief that a page is clean is below 0.36.
GRPO leaves box F1 where SFT left it (\SSftFone\ to \SGrpoFone\ over steps 50--375), trading false alarms for general recall, with brand recall at \SGrpoBrand\; on these v5 data, LoRA of rank 16 or 64 and full fine-tuning all land at \SMethodFone, and three real screens of the target hurt every critic: the bottleneck is brand knowledge, not the training method.

v5 writes the target brand and version into the prompt. Its training data have 7,822 rows from about 770 gated base pages over 34 page requirements (27\% positives), split by page, mixed with brand-free critiques the base wrote on training screenshots against forgetting.
We fine-tune the LoRA (rank 16) on the label JSON (0.7 hours on eight A100 GPUs), then continue with GRPO and serve checkpoint 150 merged into one rank-32 adapter (the served v5 model; rewards in \cref{sec:app-rl}).

\paragraph{Training methods.}
\label{sec:app-method}

All runs use the v5 data and are scored on defect test v2 (643 screenshots, 147 clean) and the general text probe (31 items).

\begin{table*}[!tp]
\centering\tstyle\setlength{\tabcolsep}{4pt}
\caption{Training methods for \judge\ on the v5 data (FT: fine-tuning; LM: language model). GRPO checkpoints from step 50 to 375 score F1 \SGrpoFone\ with brand recall \SGrpoBrand; the served model merges step 150 into one rank-32 adapter.}\label{tab:method}
\begin{tabular}{@{}llccccc@{}}
\toprule
Model & Method & False alarms & General recall & Brand recall & Defect F1 & General text \\
\midrule
\mQW & no training & 88\% & 20\% & 26\% & 0.06 & 90\% \\
\mRJ{v5} & LoRA r16 (SFT) & 37\% & 57\% & 6\% & 0.18 & 87\% \\
\mRJ{v5} & LoRA r64 & 62\% & 63\% & 12\% & 0.21 & 90\% \\
\mRJ{v5} & full FT (LM) & 65\% & 59\% & 12\% & \textbf{0.21} & 74\% \\
\mRJ{v5} & full FT (LM + vision) & 64\% & 64\% & 10\% & 0.19 & 71\% \\
\mRJ{v5} & LoRA r16 + GRPO 150 & 56\% & 58\% & 10\% & 0.19 & 90\% \\
\mRJ{v5} & same, served merged (final) & 65\% & 54\% & 11\% & 0.19 & 87\% \\
\bottomrule
\end{tabular}\end{table*}

\section{\judge: rewards and the move to a consistency check}
\label{sec:app-rl}

\begin{table*}[!tp]
\centering\tstyle\setlength{\tabcolsep}{7pt}
\caption{Critique on defect test v2 (643 screenshots, 147 clean; base pages that fail the quality gate are removed, since their clean label cannot be trusted; one prompt and one matching rule for all). \judge\ v5 is the LoRA with GRPO checkpoint 150, served as one merged adapter. Tinted: false alarms on more than half of the clean pages.}\label{tab:defect}
\begin{tabular}{@{}lcccccc@{}}
\toprule
Critic & Clean false alarms $\downarrow$ & General recall & Brand recall & Precision & F1 & Boxes / page \\
\midrule
\mGA & \lo{54\%} & 40\% & 21\% & 13\% & 0.17 & 2.0 \\
\mGF & \lo{89\%} & 32\% & 15\% & 6\% & 0.09 & 3.3 \\
\mGPT & \lo{80\%} & 47\% & 26\% & 6\% & 0.10 & 5.1 \\
\mSOL & \lo{75\%} & 50\% & \textbf{27\%} & 6\% & 0.10 & 6.0 \\
\mGP & \lo{60\%} & 25\% & 16\% & 4\% & 0.07 & 4.3 \\
\midrule
\mQW & \lo{88\%} & 20\% & 26\% & 4\% & 0.06 & 7.5 \\
\mRJ{v5, earlier, brand target} & \lo{65\%} & \textbf{54\%} & 11\% & \textbf{20\%} & \textbf{0.19} & 1.0 \\
\bottomrule
\end{tabular}
\end{table*}

\paragraph{The reward.}
Every training fault is made by the data engine, so the reward is computed by code against its labels, with no model judge in the loop. SFT teaches the answer format; GRPO then optimises a box-matching score (IoU $\geq 0.3$ or centre within 50 of 1000; the paper scores at IoU $\geq 0.5$). A parsable answer gets $+0.1$ and an unparsable one $-2.0$. On a clean page, reporting nothing gets $+1.0$, the first box costs $c_1$ and each further box $0.25$; on a faulty page the score is box F1 minus the share of faults missed, minus $0.05$ per unmatched box and minus $s$ for silence; a small bonus ($+0.1\times$ the share) rewards the right kind, and a box over half the screen matches only an equally large true box. Before any run the reward was checked against degenerate policies (report nothing, a 12-box grid, one whole-screen box, one wrong box, an unparsable answer), which must all score below the true labels. GRPO starts from the merged v5 SFT adapter and trains a new LoRA r16 on 6,004 prompts (32\% clean; 8 answers per prompt, $T$ 1.0, KL $\beta$ 0.04, lr $5{\times}10^{-6}$, 8$\times$A100-80GB).

\paragraph{Three reward versions.}
Only $c_1$ and $s$ changed. A guessed box on a page the policy believes clean with probability $p$ pays when $p < p^\ast = (s-0.05)/(c_1+s-0.05)$.
\emph{R1} ($c_1$ 1.5, $s$ 0, $p^\ast$ 0) made silence cheaper than one wrong box: clean-page false alarms fell from 37\% to 18\%, but general recall fell from \SRoneGenFrom\ to \SRoneGenTo\ and F1 from \SSftFone\ to \SRoneFoneTo. \emph{R2} ($c_1$ 1.0, $s$ 0.6, $p^\ast$ 0.36) keeps F1 at \SGrpoFone\ (\cref{tab:method}; the served step 150, merged into one r32 adapter, reaches \SFinalDtFone\ with \SFinalGenRec\ general recall), but a box pays whenever the policy is unsure: it reports about one box per page and flags almost every clean generated page. \emph{R3} ($c_1$ 1.5, $s$ 0.3, $p^\ast$ 0.14) is v6's.

\paragraph{Why generated pages.}
Two causes add up. (1) In the v5 SFT data the same clean generated picture is labelled \emph{follows} 515 times and \emph{does not follow} 1,030 times: the counterfactual rows ask about it under another brand's target. A 9B model that does not know the brands (22--23\% on the fact set, \cref{tab:knowledge}) cannot use the target, so it learns that a generated page is wrong; the SFT model already flags 57\% of them and 15\% of real screenshots. (2) R2 then rewards a guessed box whenever $p < 0.36$, and pushes that prior to its limit. Part of the 100\% may be correct: only about a third of the generated pages are recognised as their brand by the base-page gate, and every external judge also flags most of them (Gemini 4 Argon 88\%). But a critic that flags every generated page is of no use to a designer, whatever the reason.

\begin{figure}[!tbp]
\centering
\includegraphics[width=\columnwidth]{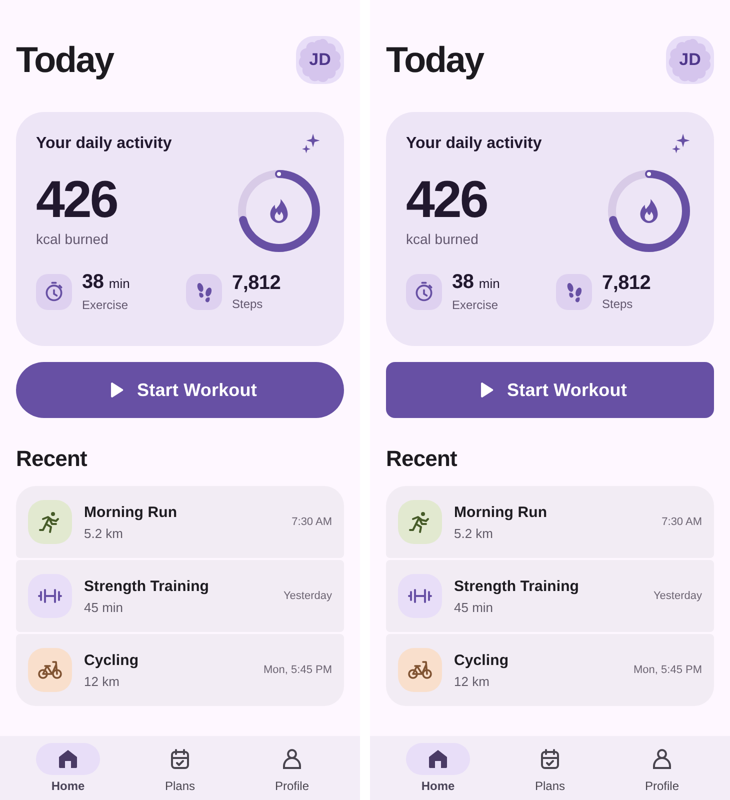}
\caption{A local style fault: the \emph{Start Workout} button's corner radius changes from 28 to 9\,px; the cards and the navigation keep the page's own shapes. The page, not a brand, is the reference.}
\label{fig:local-fault}
\end{figure}

\paragraph{v6: a consistency check, not a brand check.}
These results change the task, not only the reward. A 9B model does not learn eight brands' guidelines, but it does not need to: what goes wrong on a generated page is rarely the design itself (8--14\% of generated pages have a general defect, \cref{sec:app-brand}) and mostly one part whose style does not match the rest of the page. v6 therefore asks without a target brand: find each element that does not match the style of the rest of this page, or has a general problem. The page itself is the reference, so a clean page has one label only.
\begin{itemize}
\item \emph{Prompt.} No target line; finding kinds are \emph{common} and \emph{inconsistent} (with its aspect); the look-alike brand and the card rule leave the answer, since both need brand knowledge.
\item \emph{Data.} Out: counterfactual rows, whole-page typeface faults and old-version pages (all self-consistent, only brand knowledge tells them), and the page-wide style faults, which restyle every card or every accent and so leave the page consistent with itself. In: \emph{local} style faults. One element of a kind that the page shows at least twice takes another corner radius, material, accent or icon, and its siblings keep the page's own value. For example, one button's radius drops from 28 to 9\,px while the cards keep 28 (\cref{fig:local-fault}). Swapped components and general faults stay. The page-wide faults remain in the benchmark as brand questions for the large models.
\item \emph{Reward and selection.} R3; clean prompts repeated to 40\%; a checkpoint is usable only if its false alarms on real and on generated clean pages are both at most 15\%, and F1 decides only among usable checkpoints. No v5 model passes this gate.
\end{itemize}
The large judges already find this kind of fault without brand knowledge: on swapped components gpt-6.1-sol reaches \SSolSwapDt\ recall and gpt-6-astra \SAstraSwapDt, against \SFinalSwapDt\ for \mRJ{final}; on page-wide colour and icon faults even gpt-6.1-sol stays at \SSolStyleCI.

\paragraph{v6 results.}
\Cref{tab:v6-runs} lists the v6 runs on the consistency test (576 pages, 128 clean; 19 clean pages on which both outside judges box the same place were left out when the test was built, separately from the \BFaulty\ pages of \cref{sec:app-visibility}).
The first run (v6-a) still learned that a generated page is faulty: every faulty training image was generated, while real screenshots appeared only clean, so it flags 4\% of clean real screens but 49\% of generated ones. The second run (v6-b) repeats clean generated pages until they match the faulty ones and keeps a validation split for checkpoint choice; its chosen step 950 matches Gemini 4 Argon on false alarms but finds few local faults. The third run (v6-c) trains on full-size pages (the old pixel budget shrank 720$\times$1600 screenshots by 7\% per side), with local faults as half of the faulty rows and a GRPO bonus per found local fault; its recall carries over to the three unseen test page types, but it flags almost every clean generated test page.

\begin{table}[!tbp]
\centering\tstyle\setlength{\tabcolsep}{3pt}
\caption{\judge\ v6 runs on the consistency test. FA: clean pages flagged (real screenshots, generated pages); Swap, Local, Radius: recall. $^\dagger$Boxes kept only when $P(\text{follows}) < 0.005$; the threshold was read off the test, so this row only shows what calibration could reach, not a result.}\label{tab:v6-runs}
\resizebox{\columnwidth}{!}{\begin{tabular}{@{}lcccccc@{}}
\toprule
 & FA real & FA gen. & Swap & Local & Radius & F1 \\
\midrule
Gemini 4 Argon & 15\% & 20\% & 36\% & 32\% & 83\% & 0.31 \\
gpt-6.1-sol & 36\% & 41\% & 24\% & 28\% & 83\% & 0.20 \\
\midrule
v6-a, step 750 & 4\% & 49\% & 16\% & 5\% & 4\% & 0.22 \\
v6-b, step 950 (chosen) & 4\% & 20\% & 8\% & 7\% & 4\% & 0.21 \\
v6-b, step 1300 & 9\% & 72\% & 19\% & 13\% & 12\% & 0.24 \\
v6-c, step 2229 & 13\% & 90\% & 19\% & 21\% & 50\% & 0.25 \\
v6-c, threshold 0.005$^\dagger$ & 4\% & 13\% & -- & 16\% & -- & -- \\
\bottomrule
\end{tabular}}
\end{table}

\paragraph{Two lessons.}
(1) \emph{Validation must look like the test.} The first validation split held two page types, and on one of them (a media player with large round buttons, 67 of its 91 local faults) a changed radius is easy to see: v6-b step 950 found 15 of 24 local radius faults there but 1 of 24 on the test, although the changed elements (mostly buttons) and the size of the change (median 13--15\,px) are the same in both. v6d and v7 validate on six held-out page types.
(2) \emph{Confidence does not yet separate.} The verdict token's probability ranks clean generated pages below faulty pages only weakly (AUC 0.63 for v6-c, 0.62 for v6-b), so a threshold buys quiet with recall. v6d therefore supervised the comparison itself: each finding states what differs from its siblings (``17\,px corners; the other buttons 50\,px''), and its threshold was chosen on the validation split (six held-out page types; a later split adds 89 held-out real pages; the numbers here are on the current split). On that split v6d reaches F1 \SVsixdFone\ with \SVsixdCon\ consistency recall, but flags \SVsixdGen\ of the clean generated pages (\SVsixdReal\ of the real screenshots), so it fails the gate and is not tested.

\section{\method-Train}
\label{sec:app-train}
\method-Train is the training set the data engine makes for target-free critics, released with the benchmark (\cref{tab:train}). It holds \XTrRows\ rows on \XTrImages\ images of \XTrPages\ pages: \XTrGenPages\ generated base pages of \XTrSpecs\ page requirements that are not test requirements, and \XTrRealPages\ public web pages rendered at phone size, from which \XTrRealRows\ rows come.
\XTrClean\ of the rows are clean pages, so that a critic learns to stay silent. \XTrPairs\ designed differences pair a clean row with a faulted twin on the same element (a selected tab, a badge, a primary button, a floating button over content, a sticky bottom bar, a light placeholder): the clean row shows a difference a designer intends, the twin one that breaks the page.

The paper's \judge\ (v7f) is trained on this set (\cref{sec:app-v7}).

\section{\judge\ v7f: the paper's model}
\label{sec:app-v7}
\paragraph{v7f.}
The paper's model is trained on the v7f data with a LoRA of rank 128 ($\alpha$ 256), twice with different seeds; the two adapters are averaged and a cut is put on the verdict. Its data follow v7b's recipe (below) with three changes. (1) Validation screens leave the training set: an earlier split held rows of four validation page types, which made validation too easy; the fixed split holds six held-out page types and five more generated screens, and no training row comes from them. (2) Sixteen new training screens, none of them a test or validation screen, are drawn in eight design languages by three generators; their pages that pass the quality gate get engine faults (swaps, general and local style faults), and each of their clean pages is repeated at most four times. (3) v7b's real HTML pages (design-system templates and kit examples) are not used. In all, the \XTrRows\ rows of \method-Train (\cref{sec:app-train}; \XTrClean\ clean) and 743 auxiliary rows. Training uses v7b's settings below except the rank and the length: two epochs (2,570 steps) instead of three.
\emph{Seeds and averaging.} Two SFT runs on the same data that differ only in the seed land up to 0.04 F1 apart on validation (\cref{tab:capacity}), more than most data changes we tried. We therefore train two seeds (42 and 7) and average their adapters exactly: the update $\frac12(B_1A_1 + B_2A_2)$ is one LoRA of rank 256 with $[B_1/2, B_2/2]$ and $[A_1; A_2]$. Each run uses its final step, so validation chooses no checkpoint.
\emph{Cut.} The model's verdict gives a probability that the page is clean; a finding is kept only when that probability is below \RJCut, chosen on the validation split, first by the gate (at most 15\% of clean real and of clean generated pages flagged), then by F1.
\emph{Capacity, not data.} At rank 16, sixteen more training screens, more fault rows, GRPO and averages of up to five runs all stayed within seed noise on validation (F1 0.55--0.56). Raising the rank did not: averaged pairs reach 0.550 (rank 16), 0.571 (64), 0.607 (128) and 0.599 (256) (\cref{tab:capacity}). Each rank was chosen on validation before its test run. On the v5 data, rank 16 and 64 had scored alike (\cref{tab:method}); with the v7f data, the adapter size became the limit. Three epochs instead of two (averaged pair \CapThreeEp, +0.012) stay within the noise and are not used.
\emph{Votes over runs.} Instead of averaging weights, four rank-128 runs (two seeds on the v7f data, two on the v7h data, which adds sixteen generated screens) can vote: a finding stays when at least three runs box the same place (IoU $\geq 0.5$), and the cut is put on their mean verdict probability. Fixed on validation (F1 \CapVoteVal\ against \CapPairVal\ for the paper's model) and tested once, the vote raises F1 on the main test from \RJFone\ to \CapVoteTest\ (paired bootstrap \CapVoteDiff, 95\% interval $[\CapVoteLo, \CapVoteHi]$) at \CapVoteRealFA\ and \CapVoteGenFA\ flagged clean real and generated pages, but it flags \CapVoteOldGen\ of the clean generated pages of the first 576-page test and \CapVoteText\ of the clean text layout pages, beyond the gate, and falls on OwlEye (F1 \CapVoteOw\ against \XOwFone), at four answers per page. We therefore keep the single averaged model, although the vote is higher on validation.
\emph{Every test run.} The main test was scored for \CapRuns\ variants of the v7 line (\cref{tab:capacity}): v7e with GRPO (F1 \CapVeGrpo), v7f at rank 16 (\RJPrevFone, the paper's model before this one) and with GRPO (\CapVfGrpo), the v7h pair (\CapVhPair), the rank-64 pair (\CapSixtyFour), the rank-128 pair (\RJFone, ours) and the vote (\CapVoteTest). Each was fixed on validation before its test run, but the choice among the tested variants saw their test scores. Two rank-128 variants score slightly higher on validation, within seed noise, and were not tested (three epochs \CapThreeEp, the v7h data \CapVhVal).
\emph{Test.} On the \RJPages-page consistency test (\cref{tab:consistency}) it reaches F1 \RJFone\ and flags \RJRealFA\ of clean real and \RJGenFA\ of clean generated pages; without the cut, F1 \RJRawFone\ with \RJRawRealFA\ and \RJRawGenFA. On the first 576 pages of the test, used for the earlier runs below, it reaches F1 \RJOldFone\ (Gemini 4 Argon \RJOldArgonFone) and flags \RJOldRealFA\ and \RJOldGenFA.

\paragraph{v7b, the earlier model.}
\label{sec:app-v7b}
v7 keeps v6's target-free prompt, with a short \emph{why} field per finding, and changes the data in three ways. Clean generated training pictures that Gemini 3.8 Flash and Gemini 4 Argon both box in the same place are dropped (500 of 1,831 pictures, 928 rows; 10 of 75 clean generated test pages are disputed in the same way). New copies carry one to four independent faults of different kinds with boxes that do not overlap, each checked to change visible pixels; besides overflow, overlap, small targets, low contrast and the local style faults, the engine makes display faults of the kinds UI-Lens and OwlEye label (a null value, a missing image, a cropped element, a wrong ellipsis) in our own page code, and no UI-Lens or OwlEye page is used. 19 real HTML pages (design-system templates and kit examples) join as faulted copies and as clean pages. In all, 11,544 rows from 543 pages, 43\% clean; faulted rows hold one (3,772), two (1,330), three (909) or four (582) faults.
v7b is LoRA r16 ($\alpha$ 32), lr $10^{-4}$, batch 14 on seven A100 GPUs, 3 epochs (2,646 steps). Step 1600 was chosen on six held-out page types under an earlier match rule that also counted a box whose centre lay inside the true box grown by 5\% of the screen; rescored with the main rule, its validation F1 is \SValFone, and the best checkpoint that passes the gate (step \SValBestStep, F1 \SValBestFone, \SValBestGenFA\ of the clean generated pages) scores \SValBestTest\ on the first 576 test pages. On the main test v7b reaches F1 \RJVbFone\ and flags \RJVbRealFA\ of clean real and \RJVbGenFA\ of clean generated pages (on the first 576 pages, F1 \SValPrevTest; v6: \XVsixFone).

\paragraph{A failed run: clean real screens only.}
v7c adds 2,592 real screenshots (ColorBench, iOS-1K, MMBench-GUI and web pages) to the data of v7, all labelled clean, and trains for two epochs. Validation picks step 650 by the gate (\SVsevencValReal\ of clean real and \SVsevencValGen\ of clean generated pages flagged), but the model has learned that a real screenshot is clean: it flags \SVsevencOwRec\ of OwlEye's faulty screens and reaches consistency recall \SVsevencCon\ and F1 \SVsevencFone\ on the first 576 pages of the consistency test. No real screen in its training data carried a fault, so real screens need faulted copies as well as clean ones.

\section{Find, then fix}
\label{sec:app-fix}

\begin{table*}[!t]
\centering\tstyle\setlength{\tabcolsep}{3pt}
\caption{Find, then fix: gpt-6.1-sol repairs 85 pages with one local fault and is also given 40 clean generated pages. Sent: pages that reach the fixer. ours: the paper's model; r16 + cut: the same data with a rank-16 LoRA (the previous model); v7b: an earlier model; v6-c and v6-b: earlier one-fault runs.}\label{tab:fix}
\begin{tabular}{@{}lccccccc@{}}
\toprule
Hint & Fixed & Radius & Colour & Material & Harm, faulty & Harm, clean & Sent \\
\midrule
None & 95\% & 100\% & 69\% & 100\% & 0\% & 12\% & 100\% \\
True finding & 95\% & 100\% & 77\% & 98\% & 0\% & 0\% & 100\% \\
\midrule
Gemini 4 Argon findings & 55\% & 79\% & 38\% & 48\% & 11\% & 15\% & 100\% \\
\judge\ (ours) findings & 25\% & 29\% & 8\% & 27\% & 1\% & 0\% & 100\% \\
\judge\ r16 + cut (earlier) findings & 16\% & 8\% & 8\% & 23\% & 0\% & 2\% & 100\% \\
\judge\ v7b (earlier) findings & 42\% & 67\% & 8\% & 40\% & 6\% & 8\% & 100\% \\
\judge\ v6-c (earlier) findings & 39\% & 46\% & 8\% & 44\% & 8\% & 8\% & 100\% \\
\judge\ v6-b (earlier) findings & 13\% & 4\% & 8\% & 19\% & 2\% & 0\% & 100\% \\
\midrule
Gate: Gemini 4 Argon & 72\% & 92\% & 38\% & 71\% & 0\% & 10\% & 56\% \\
Gate: \judge\ (ours) & 28\% & 38\% & 15\% & 27\% & 0\% & 5\% & 22\% \\
Gate: \judge\ r16 + cut (earlier) & 21\% & 17\% & 15\% & 25\% & 0\% & 0\% & 17\% \\
Gate: \judge\ v7b (earlier) & 72\% & 88\% & 23\% & 77\% & 0\% & 12\% & 64\% \\
Gate: \judge\ v6-c (earlier) & 94\% & 100\% & 62\% & 100\% & 0\% & 12\% & 96\% \\
Gate: \judge\ v6-b (earlier) & 24\% & 12\% & 23\% & 29\% & 0\% & 2\% & 19\% \\
\bottomrule
\end{tabular}
\end{table*}

Does a judge's finding help a generator repair the page? We test it where the repair can be checked exactly.

\paragraph{Pages.} 85 consistency-test pages with one local fault whose right value the page itself shows (24 corner radius, 13 accent colour, 48 surface material: one element differs from its siblings), and 40 clean generated pages, on which a false finding should change nothing. Icon and swap faults are left out: the original icon or component cannot be known from the page, so agreement with the clean original would be the wrong measure. The engine keeps screenshots only, so each faulty page is rebuilt from its record and kept only if its render matches the stored screenshot (125 of 128). The fault is written as an ordinary rule in the page's own stylesheet (a structural selector at a random place among the rules), so that the code does not give it away. A first run that left it as an inline \texttt{!important} style is discarded: that style gave the fault away, and the fixer found it with no hint as often as with the true finding.

\paragraph{Fixer and hints.} gpt-6.1-sol (Codex, medium effort) gets the page's HTML, its screenshot and one hint: \emph{none} (fix whatever does not match the rest of the page, or return it unchanged); the \emph{true finding} (box, kind, aspect and what differs, e.g.\ ``7\,px corners; the other buttons 20\,px''); or a judge's findings on the consistency test, with the instruction to fix the reported problems that are real and change nothing else. A judge that reports nothing leaves the page as it is. As a \emph{gate}, the judge only decides whether a page goes to the fixer, which then works with no hint.

\paragraph{Scores.} Against the clean original rendered the same way: a fault is \emph{fixed} when the pixels inside its box that differ from the clean page fall to a quarter or less; \emph{harm} is a change of more than 1\% of the pixels outside the box (faulty pages) or anywhere (clean pages).

\paragraph{Reading.} Repair is not the bottleneck: with the page's code and screenshot, the fixer finds and fixes 95\% of these faults unprompted and never damages the rest of a faulty page. Run on every page, however, it also changes 12\% of clean pages. Only the true finding gives both a high fix rate and no harm. A judge's findings help only as far as they are right: under ``change nothing else'' a finding on the wrong element keeps the fixer from the real fault, so even Gemini 4 Argon's findings lower the fix rate from 95\% to 55\%. As a gate, the paper's \judge\ sends only 22\% of the pages and lets 28\% of the faults be fixed, harming 5\% of clean pages (the rank-16 model: 17\%, 21\%, none), while Gemini 4 Argon sends 56\% and 72\% are fixed; the earlier v7b sends 64\% (72\% fixed); the earlier v6-c sends 96\% of the pages, and v6-b protects clean pages (2\% harmed) but lets only 24\% of the faults through.

\paragraph{Limits.} The faults are injected, and a stray rule with a long structural selector may still be easier to spot in code than an inconsistency a generator writes itself. Without code (a screenshot, a design-tool frame or an installed app) the fixer cannot read the page, and the judge's location matters more; we did not test that setting. Each condition ran once.

\section{More detection examples}
\label{sec:app-qual}
\begin{figure*}[tp]
\centering
\includegraphics[width=0.64\textwidth]{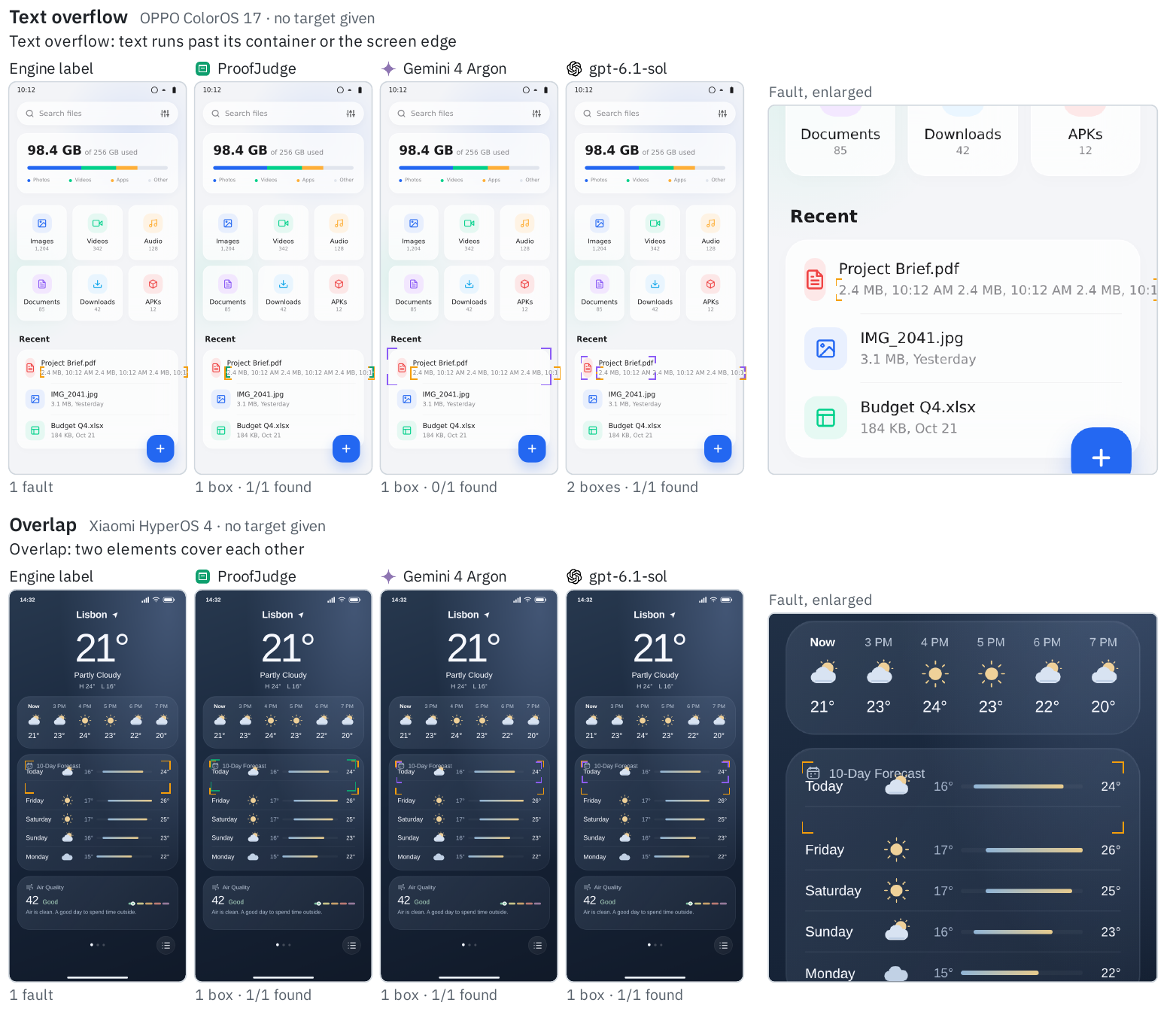}
\caption{General faults on the consistency test (no target given, IoU $\ge 0.5$). Text overflow: found by {\judge}, gpt-6.1-sol; Overlap: found by {\judge}, Gemini 4 Argon, gpt-6.1-sol.}
\label{fig:qual-general1}
\end{figure*}

\begin{figure*}[tp]
\centering
\includegraphics[width=0.64\textwidth]{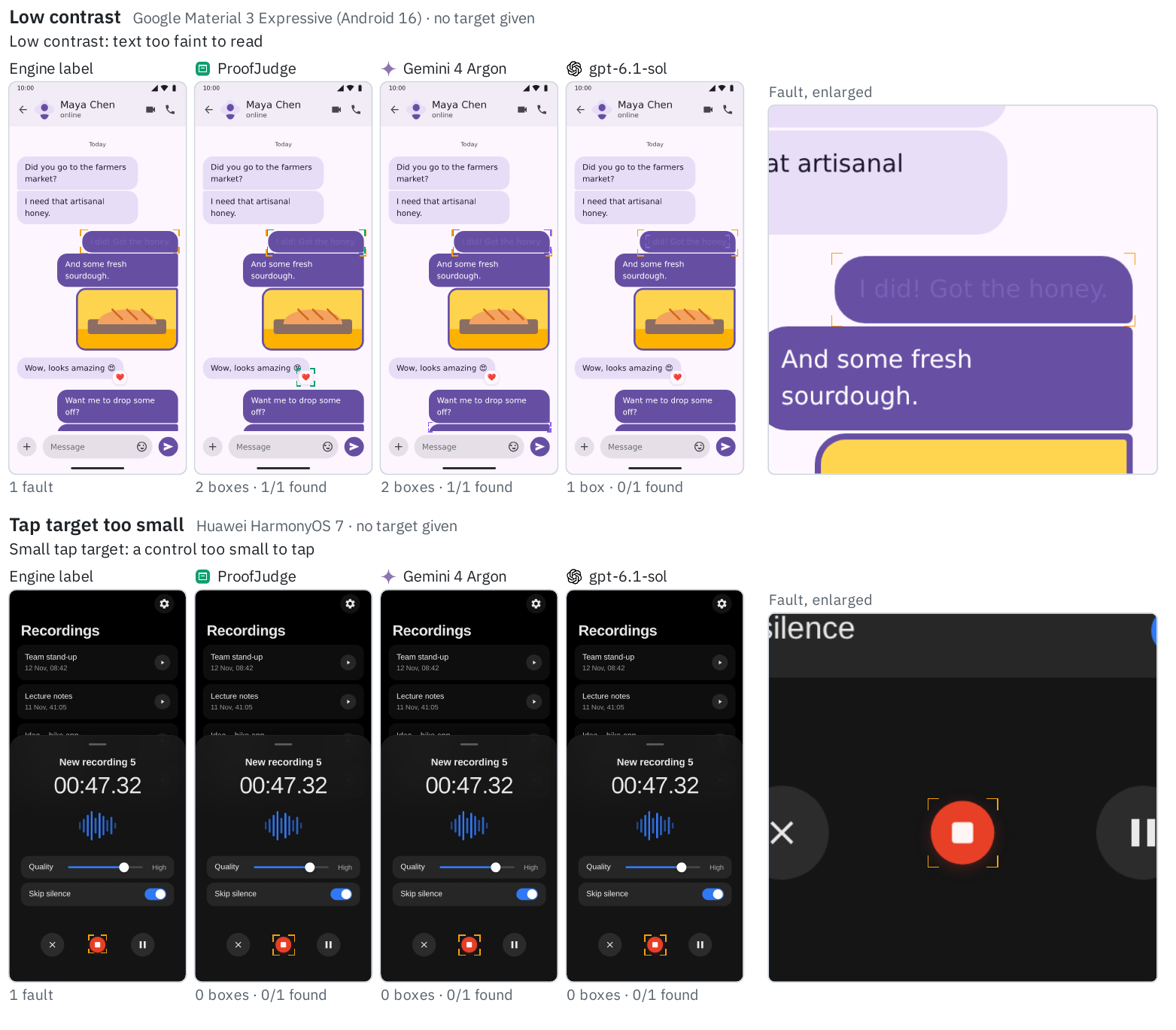}
\caption{General faults on the consistency test (no target given, IoU $\ge 0.5$). Low contrast: found by {\judge}, Gemini 4 Argon; Tap target too small: found by no judge.}
\label{fig:qual-general2}
\end{figure*}

\begin{figure*}[tp]
\centering
\includegraphics[width=0.64\textwidth]{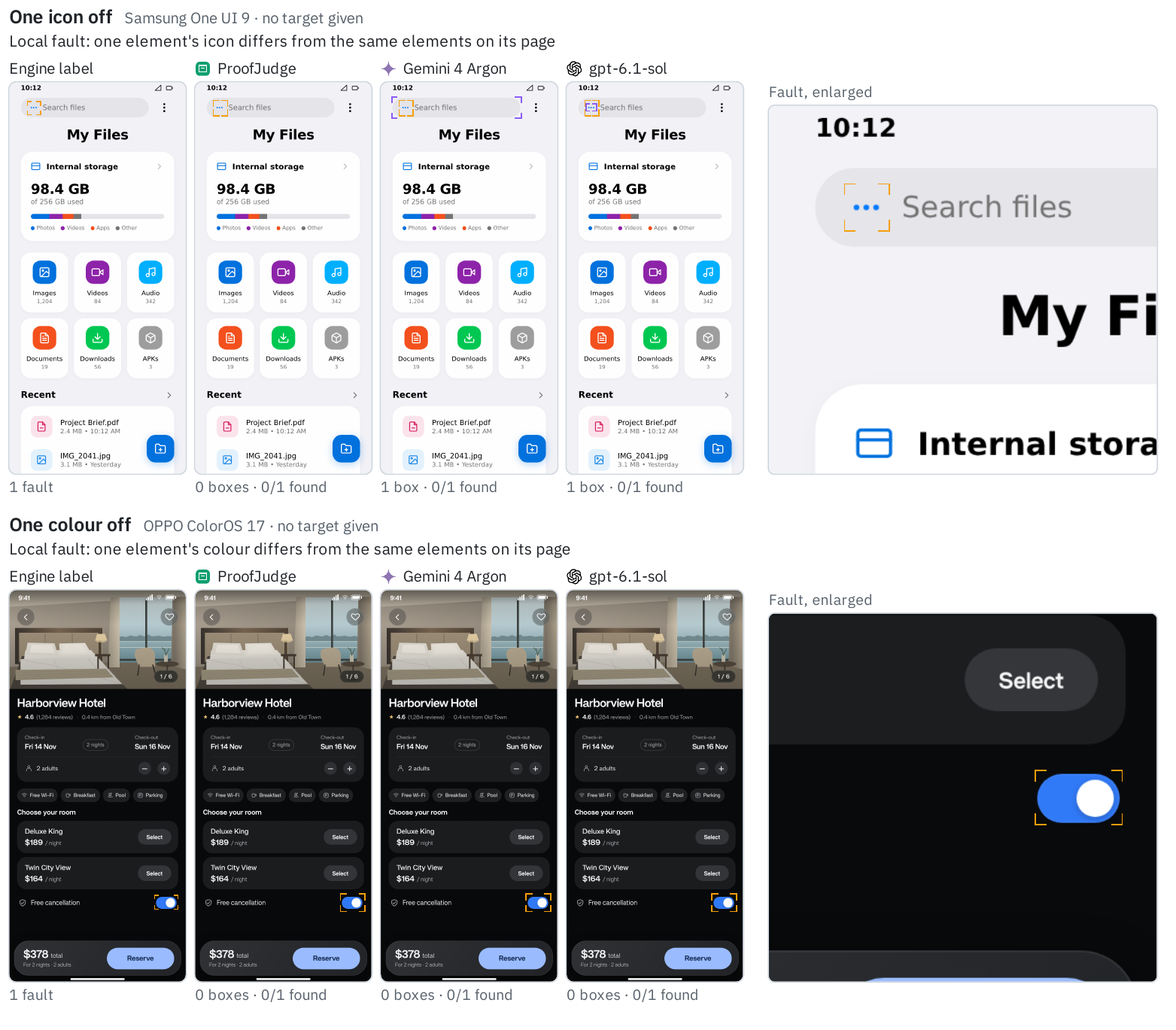}
\caption{Local style faults on the consistency test (no target given, IoU $\ge 0.5$). One icon off: found by no judge; One colour off: found by no judge.}
\label{fig:qual-local1}
\end{figure*}

\begin{figure*}[tp]
\centering
\includegraphics[width=0.64\textwidth]{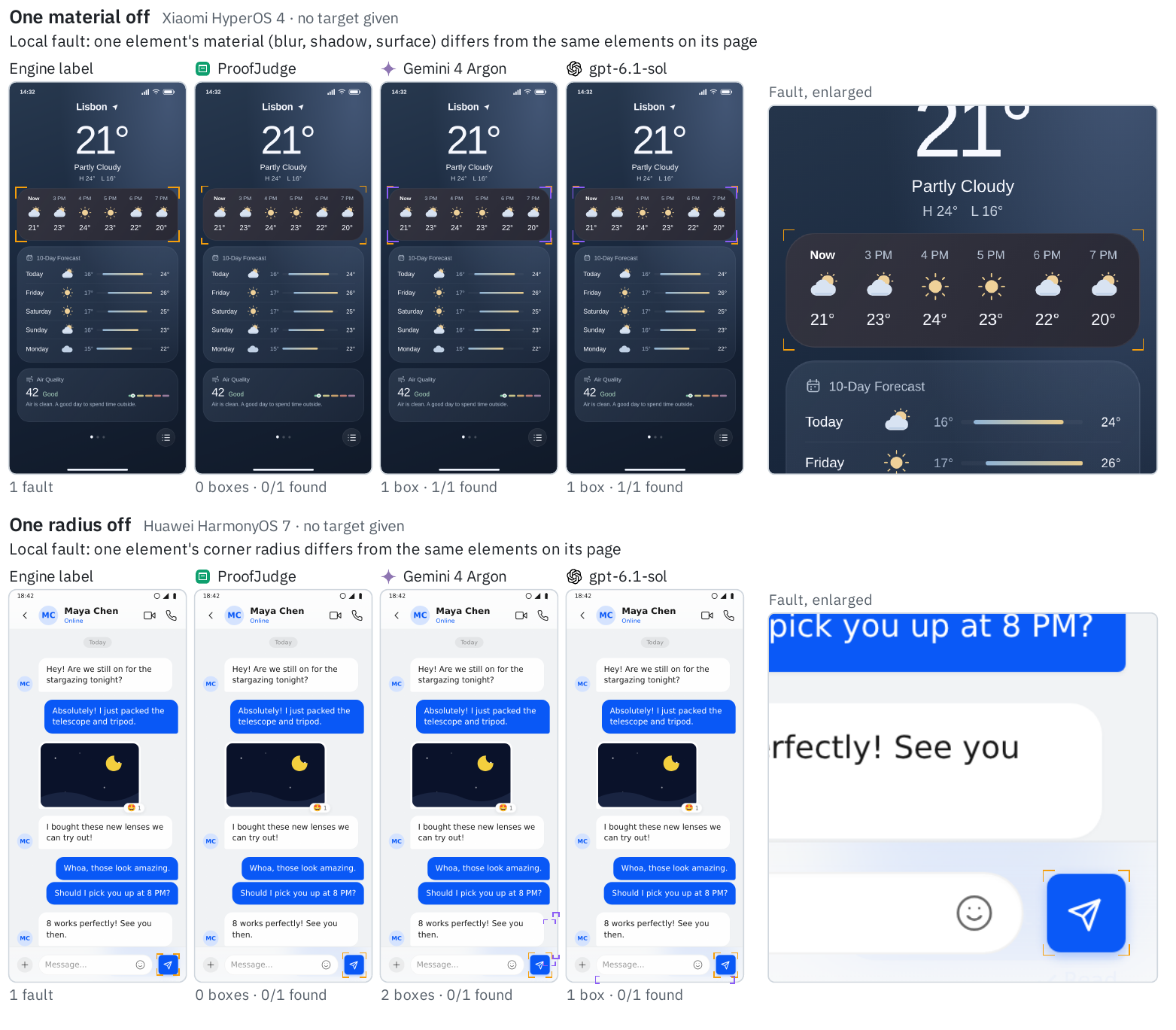}
\caption{Local style faults on the consistency test (no target given, IoU $\ge 0.5$). One material off: found by Gemini 4 Argon, gpt-6.1-sol; One radius off: found by no judge.}
\label{fig:qual-local2}
\end{figure*}

\begin{figure*}[tp]
\centering
\includegraphics[width=0.64\textwidth]{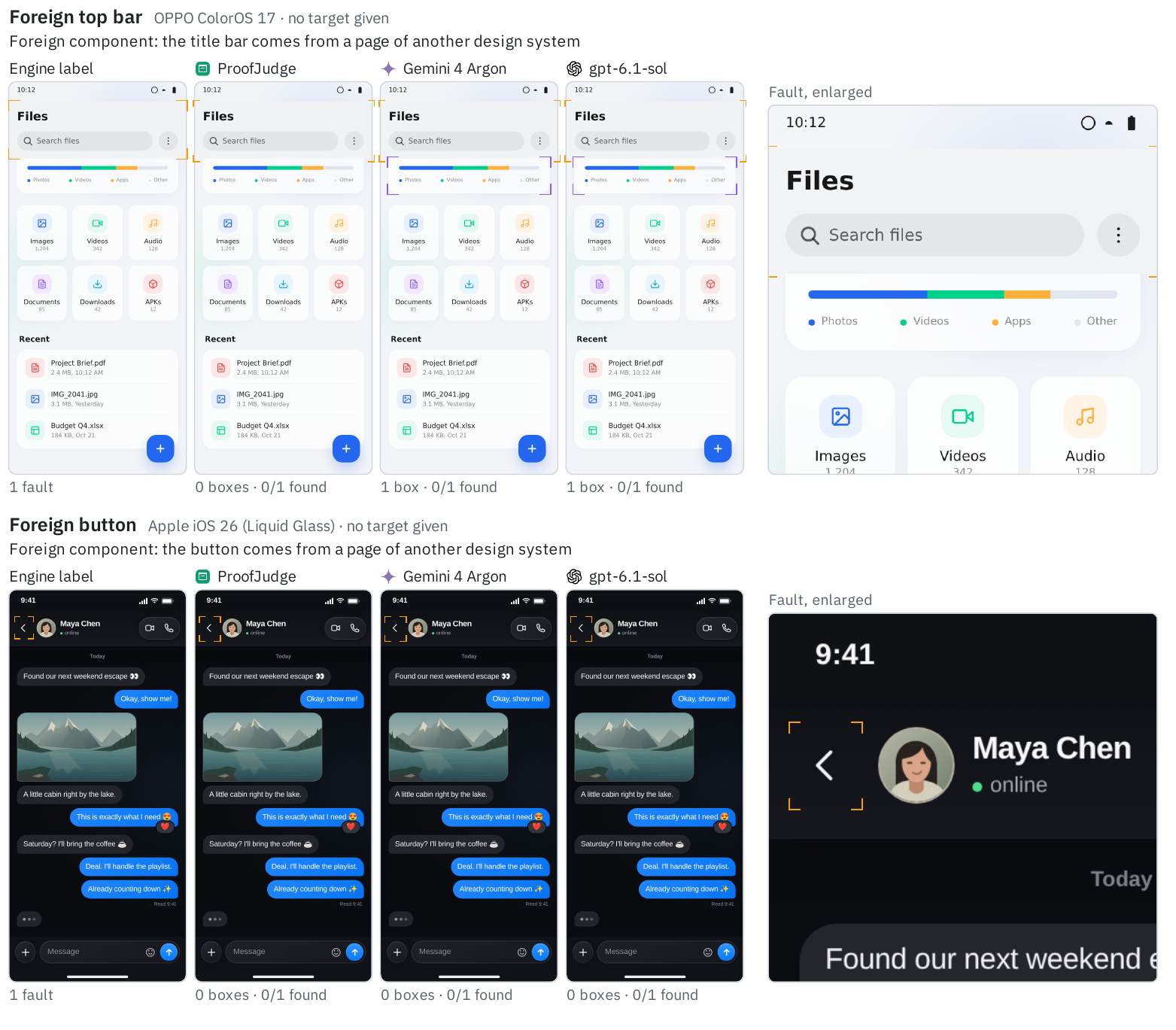}
\caption{Components from another design system on the consistency test (no target given, IoU $\ge 0.5$). Foreign top bar: found by no judge; Foreign button: found by no judge.}
\label{fig:qual-component1}
\end{figure*}

\begin{figure*}[tp]
\centering
\includegraphics[width=0.64\textwidth]{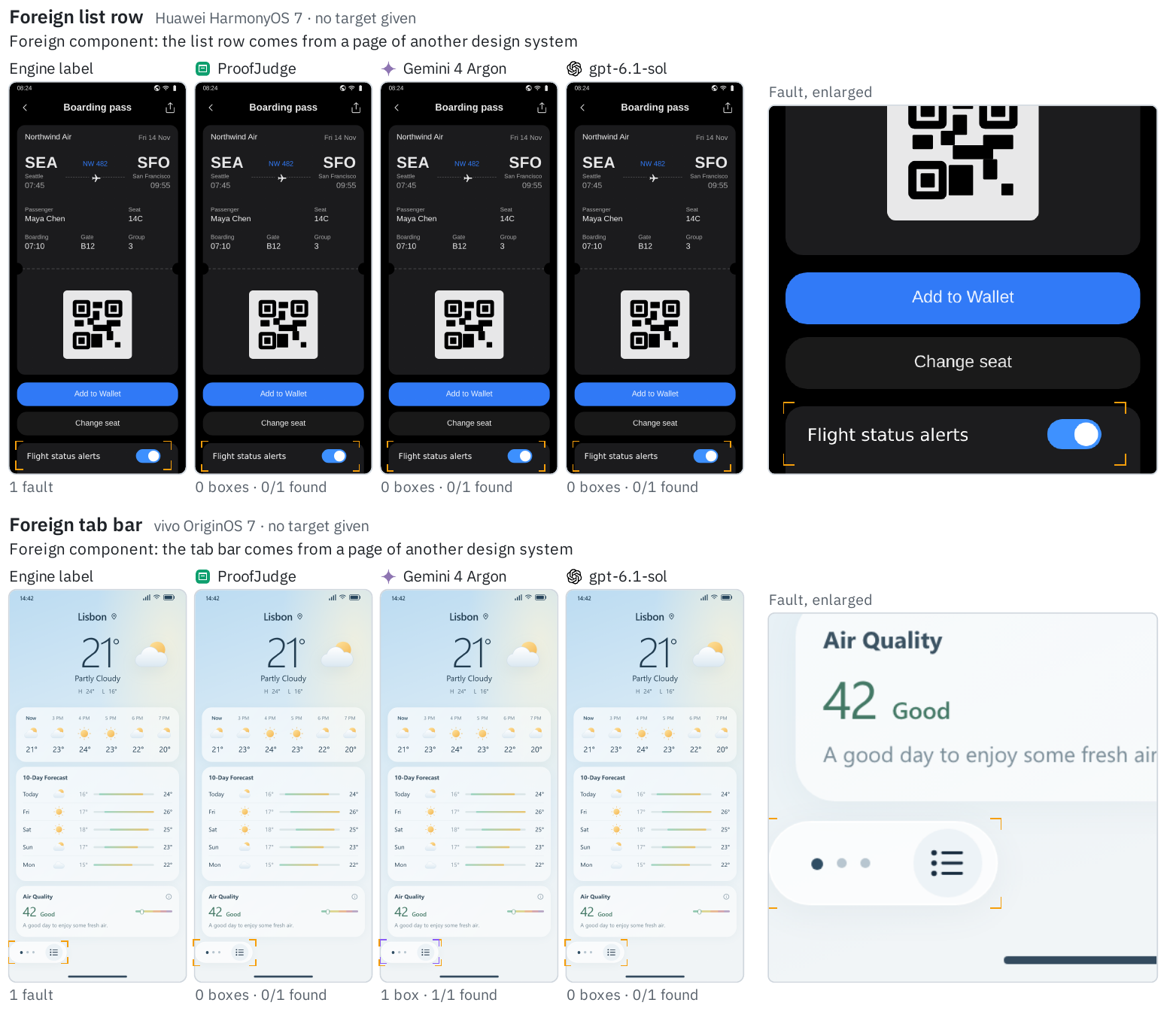}
\caption{Components from another design system on the consistency test (no target given, IoU $\ge 0.5$). Foreign list row: found by no judge; Foreign tab bar: found by Gemini 4 Argon.}
\label{fig:qual-component2}
\end{figure*}

\begin{figure*}[tp]
\centering
\includegraphics[width=0.64\textwidth]{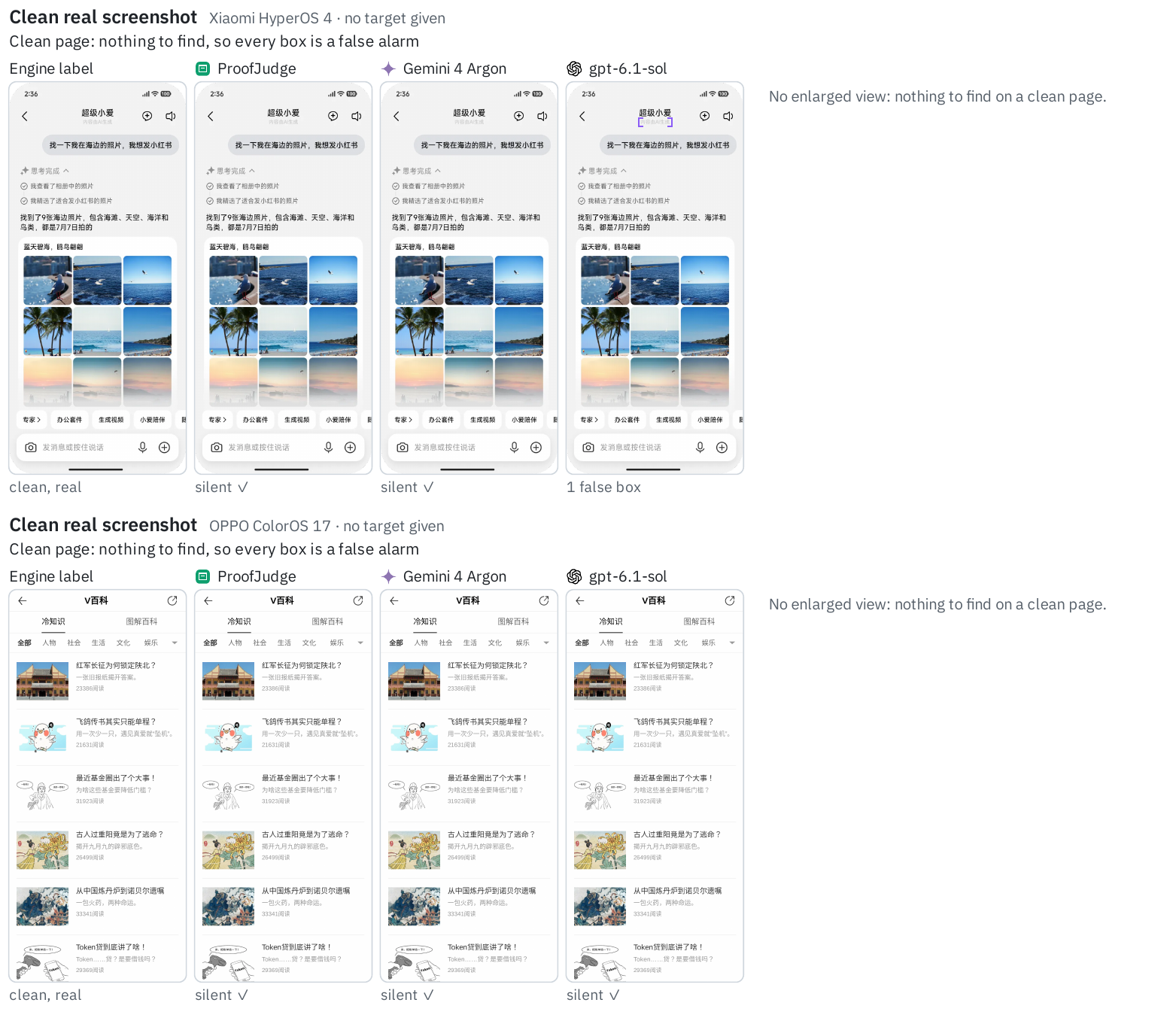}
\caption{Clean pages on the consistency test (no target given, IoU $\ge 0.5$). Clean real screenshot: false boxes from gpt-6.1-sol (1); Clean real screenshot: every judge stays silent.}
\label{fig:qual-clean1}
\end{figure*}

\begin{figure*}[tp]
\centering
\includegraphics[width=0.64\textwidth]{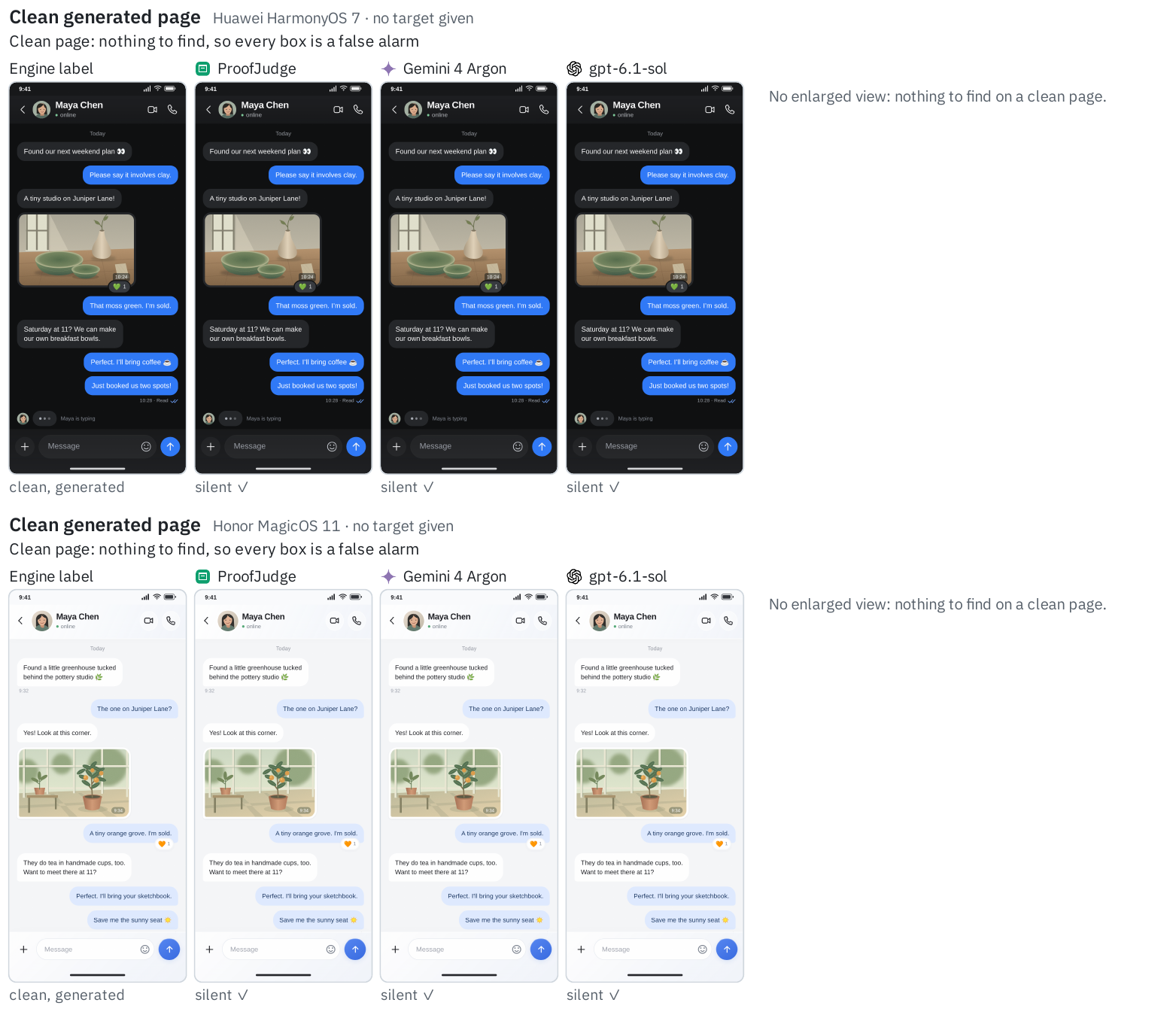}
\caption{Clean pages on the consistency test (no target given, IoU $\ge 0.5$). Clean generated page: every judge stays silent; Clean generated page: every judge stays silent.}
\label{fig:qual-clean2}
\end{figure*}

\begin{figure*}[tp]
\centering
\includegraphics[width=0.64\textwidth]{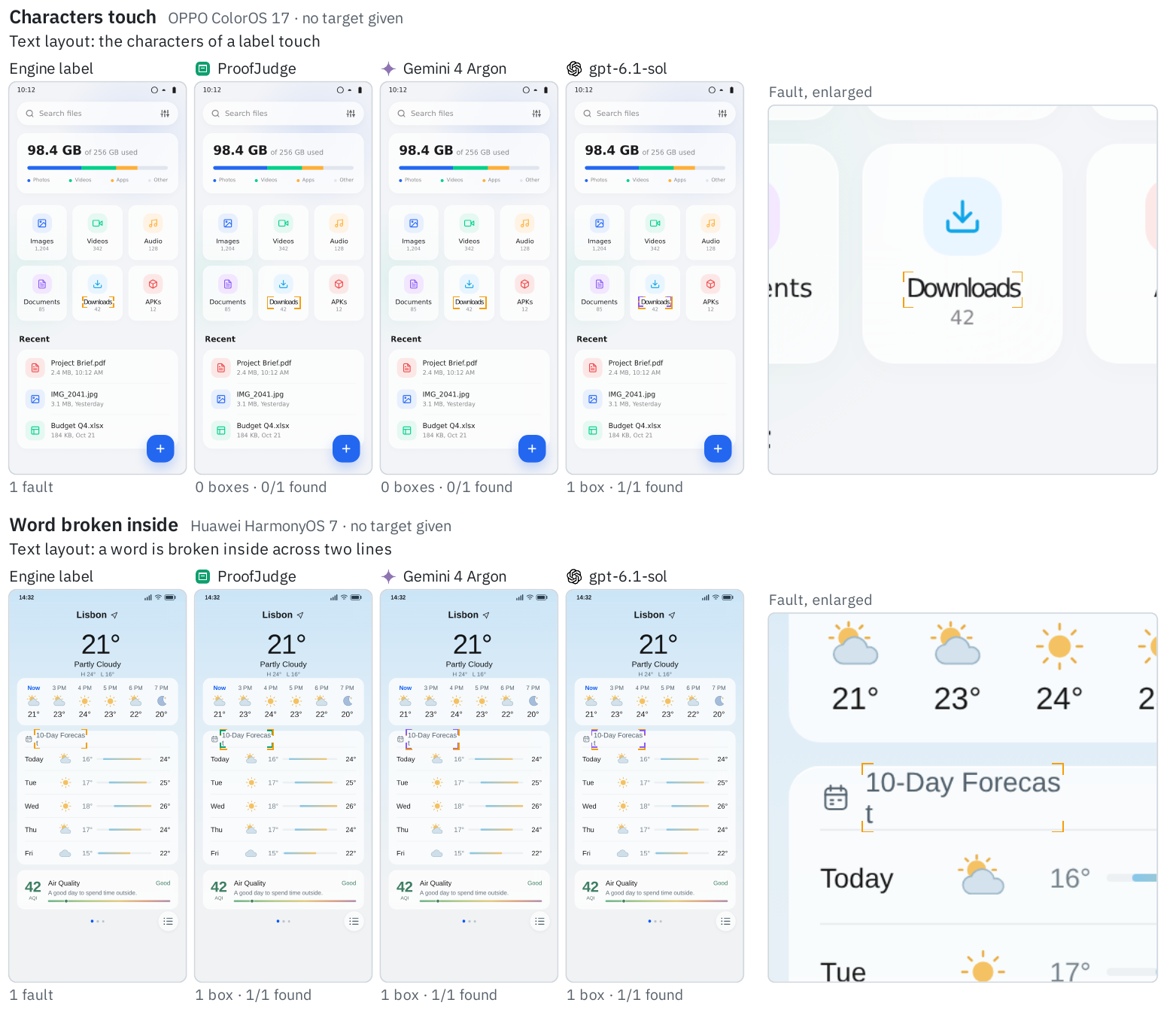}
\caption{Text layout faults on the text layout test (no target given, IoU $\ge 0.5$). Characters touch: found by gpt-6.1-sol; Word broken inside: found by {\judge}, Gemini 4 Argon, gpt-6.1-sol.}
\label{fig:qual-text1}
\end{figure*}

\begin{figure*}[tp]
\centering
\includegraphics[width=0.64\textwidth]{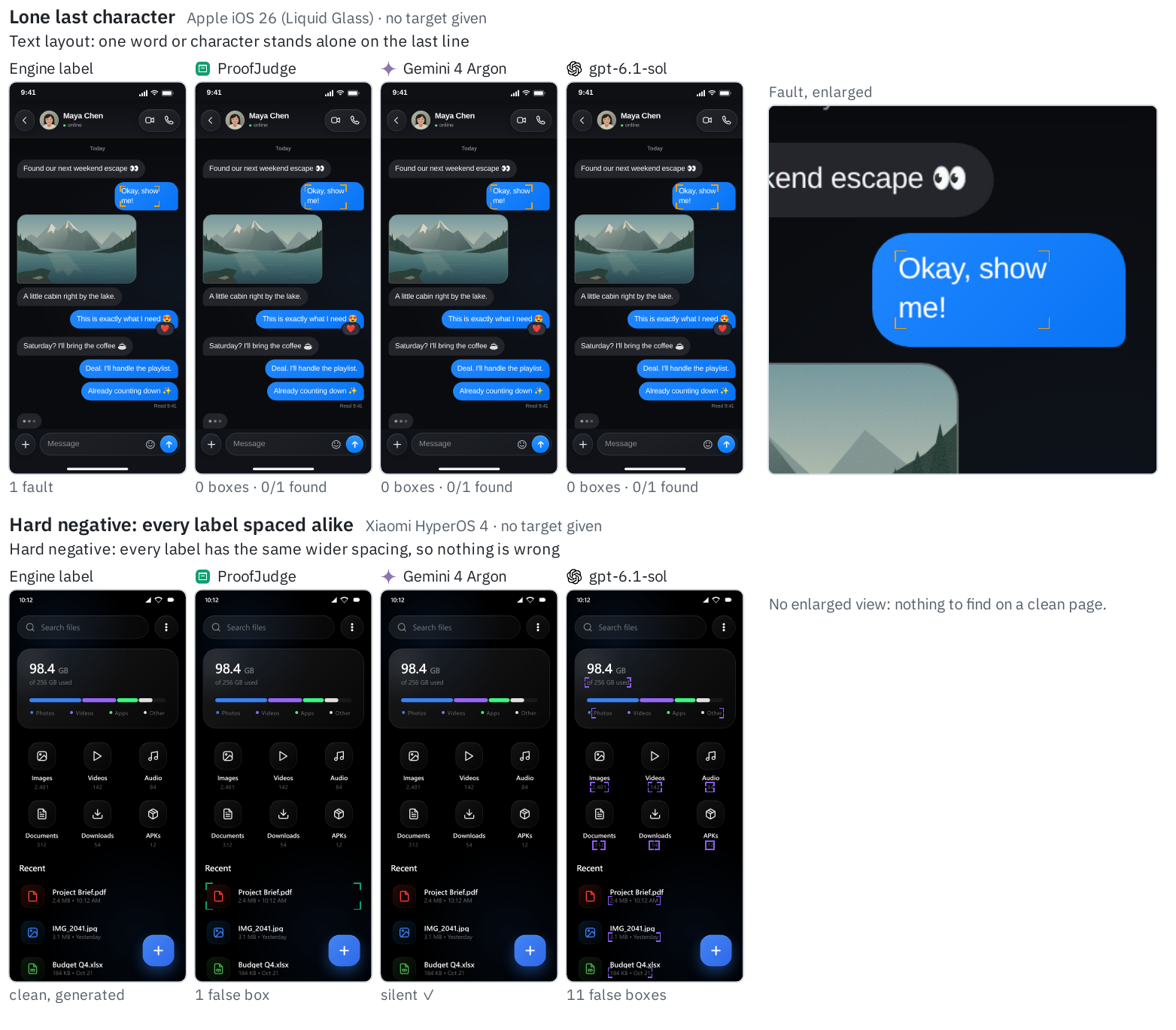}
\caption{Text layout faults on the text layout test (no target given, IoU $\ge 0.5$). Lone last character: found by no judge; Hard negative: every label spaced alike: false boxes from {\judge} (1), gpt-6.1-sol (11).}
\label{fig:qual-text2}
\end{figure*}

\Cref{fig:qual-general1,fig:qual-general2,fig:qual-local1,fig:qual-local2,fig:qual-component1,fig:qual-component2,fig:qual-clean1,fig:qual-clean2,fig:qual-text1,fig:qual-text2} extend \cref{fig:qual} with twenty more test screenshots (\judge\ v7f, no target): four of each kind on the consistency test (general faults, local style faults, components from another design system, clean pages) and four on the text layout test.
Unlike the hand-picked cases of the main figure, every case here is the first test page of its kind, with one brand and one page per case where possible and no condition on any model's answer, so successes and failures appear in their real proportion.
Two kinds of page are skipped, because a judge's box there may mark a real problem the label does not hold: clean pages that are disputed (two API judges box one place) or that the blind check found faulty (\cref{sec:app-visibility}), and every page made from a base page with such a problem, or on any copy of which both API judges box one place that touches no labelled fault (for example a send button cut off at the screen edge).
Each case names its fault in one line and enlarges the faulty area on the right; every box is drawn as four corner marks just outside its region: amber marks are the engine's exact label, mint marks \judge's boxes and violet marks the API critics'. The captions state each case's outcome; figures and captions are generated by code from the stored answers.

\FloatBarrier

\end{document}